\documentclass{article}

\usepackage{microtype}
\usepackage{graphicx}
\usepackage{subfigure}
\usepackage{booktabs} 

\usepackage{hyperref}

\usepackage[accepted]{icml2023}

\usepackage{amsmath}
\usepackage{amssymb}
\usepackage{mathtools}
\usepackage{amsthm}

\usepackage{arydshln} 
\usepackage{multirow} 
\usepackage{wrapfig} 
\usepackage{tabularx} 
\usepackage{booktabs}
\usepackage{float}
\usepackage{placeins}

\usepackage[capitalize,noabbrev]{cleveref}

\theoremstyle{plain}

\theoremstyle{definition}

\theoremstyle{remark}

\usepackage[textsize=tiny]{todonotes}

\icmltitlerunning{Sparsified Model Zoo Twins: Investigating Populations of Sparsified Neural Network Models}

\begin{document}

\twocolumn[
\icmltitle{Sparsified Model Zoo Twins: \\Investigating Populations of Sparsified Neural Network Models}




\begin{icmlauthorlist}
\icmlauthor{Dominik Honegger}{yyy}
\icmlauthor{Konstantin Schürholt}{yyy}
\icmlauthor{Damian Borth}{yyy}
\end{icmlauthorlist}

\icmlaffiliation{yyy}{AIML Lab, School of Computer Science, University of St.Gallen, St.Gallen, Switzerland}

\icmlcorrespondingauthor{Konstantin Schürholt}{konstantin.schuerholt@unisg.ch}

\icmlkeywords{ModelZoo, Sparsification}

\vskip 0.3in
]



\printAffiliationsAndNotice{}  

%
%
\begin{abstract}
With growing size of Neural Networks (NNs), model sparsification to reduce the computational cost and memory demand for model inference has become of vital interest for both research and production.
While many sparsification methods have been proposed and successfully applied on individual models, to the best of our knowledge their behavior and robustness has not yet been studied on large populations of models.
With this paper, we address that gap by applying two popular sparsification methods on populations of models (so called model zoos) to create sparsified versions of the original zoos.
We investigate the performance of these two methods for each zoo, compare sparsification layer-wise, and analyse agreement between original and sparsified populations.
We find both methods to be very robust with magnitude pruning able outperform variational dropout with the exception of high sparsification ratios above 80\%.
Further, we find sparsified models agree to a high degree with their original non-sparsified counterpart, and that the performance of original and sparsified model is highly correlated. 
Finally, all models of the model zoos and their sparsified model twins are publicly available: \url{modelzoos.cc}.
\end{abstract}

%
%
%

\section{Introduction}
In recent years, deep neural networks have gained significant momentum and popularity 
with the general trend of growing in size. This is mainly due to the observed relationship
between model size and performance i.e. larger models tend to have an improved performance
over their smaller counterparts as reported by~\cite{kaplan2020scaling,tan2019size,brock2018scale}.
%
%
Unfortunately, the increasing performance results in very high computational and environmental 
costs for training and inference, as the size of the models continuous to increases~\cite{hoefler2021sparsity,strubell2019energy}.
As an example, the image classification model CoCa, which currently achieves the highest accuracy 
(91.0\%) on the ImageNet dataset, has 2.1 billion parameters \cite{yu2022coca}. Forecasts predict 
that by 2025 models will exist able to achieve a performance of 95\% on the ImageNet object 
classification but demand as much electricity for training as New York City emits in $CO_2$ in a 
month~\cite{thompson2022}. 
\looseness-1

\begin{figure*}[] 
\centering
{\includegraphics[width = 1.8\columnwidth]{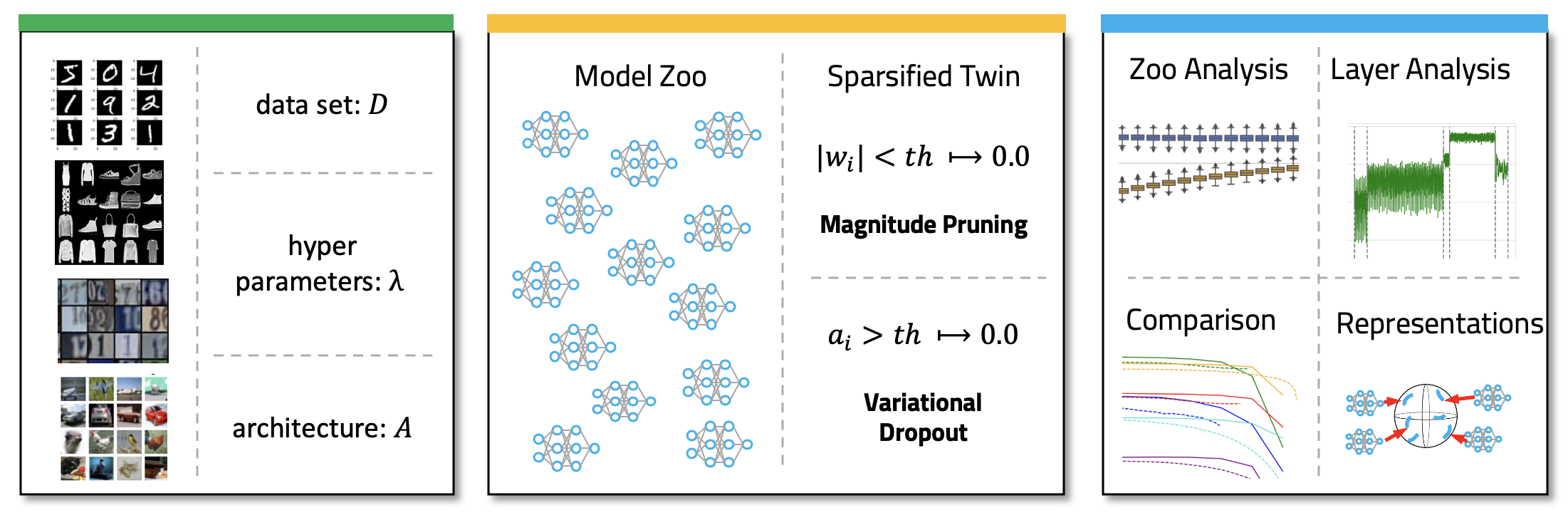}}
\vspace{-0.25cm}
\caption[Approach]{An overview of the approach. (Left:) A population of neural network models is trained according to latent generating factors such as dataset, architecture and hyperparameter. (Middle:) The given model zoos are sparsified given magnitude pruning and variational dropout. (Right:) Models in the populations are analyzed and entire model zoos are compared with each other and its sparsified counterpart. Additional representation of the zoos are trained to analyse the underlying structure of the sparsified zoos.}
\label{fig:pipeline}
\end{figure*}

One approach to tackle this issue is to exploit the over-parameterization of large models of neural 
networks (NNs) to successfully train, but to reduce their size significantly for inference. According to~\cite{hoefler2021sparsity} 
sparsification of neural network models can achieve reductions of 10-100x without significant losses in 
performance, even for extremely large models \citep{sparsegpt2023frantar}. By pruning parameters after 
training, it becomes possible to reduce the required computational power for inference, save energy or 
deploy models on mobile devices, on embedded systems or satellites with limited storage capabilities~\cite{giuffrida2021varphi, hoefler2021sparsity, howard2017mobilenets}. 

Related work has investigated individual methods of sparsification extensively~\cite{blalock2020state, hoefler2021sparsity} 
and run large scale studies rigorously evaluating performance differences between different methods~\cite{gale2019state}. 
Contrary to~\cite{gale2019state}, who evaluate sparsification on fixed seeds and optimize hyper-parameters for best sparsification, this 
work evaluates the effect on specification on populations of neural network models (so called ``model zoos''). 
Since neural networks follow a non-convex optimization and are sensitive to hyper-parameter selection, 
to achieve more robust results in studying sparsity we 
propose to shift the focus from individual models to populations of neural networks~\cite{schuerholt2021ssl, schurholt2022gener}, 
which are trained according to controlled generating factors i.e., selection 
of hyper-parameters, seeds, initialization methods. To the best of our knowledge, there are no studies 
on sparsification on a population of neural network models.

\paragraph{Our contributions:} 
(1) We generate a sparsified version of an available model zoo~\cite{schurholt2022model} 
using two popular sparsification methods, namely Variational Dropout (VD)~\cite{molchanov2017variational} 
and Magnitude Pruning (MP)~\cite{han2015deep,strom1997sparse} and thus generate a dataset consisting of 33'920 
trained and sparsified CNNs with 1'721'600 unique model states representing their sparsification trajectories. 
(2) We conduct an in-depth analysis and comparison of the sparsified model zoos and the utilized sparsification 
methods and find that i) both methods perform robustly on all populations, ii) MP outperforms VD except for 
some very high sparsity ratios and iii) higher sparsity ratios are achieved in larger layers consistently in 
the populations of the model zoos. Since for each individual model a dense and fully parameterised as well 
as a sparsified version exists, their relationships can be investigated. Particular attention is paid to 
investigating how robustly the methods perform on the model zoos trained on different datasets and with 
varying hyperparameter configurations. 
(3) As expected, on average with increased sparsification, we observe a performance drop in the populations. 
However, within the population, we can find individual models, which are less prone to the performance drop 
(they are sparsification-friendly) or vice verse, are affected stronger by the performance drop (they are 
sparsification-hard).
(4) Furthermore, we examine the weight spaces of the sparsified model zoos by learning hyper-representations 
of the individual model parameters and are able to show that model properties such as accuracy and sparsity 
disentangle very well in the latent space and can be predicted from its latent representation.\looseness-1

%
%
\section{Related Work}
\paragraph{Sparsification of Neural Networks} Model sparsification has been studied in depth, \cite{hoefler2021sparsity} provides a survey over the different approaches. 
Most sparsification approaches can be categorized as 'data-free' or 'training-aware'. 
Data-free approaches prune models based on the structure of the neural networks. Magnitude Pruning (MP) \cite{han2015deep,strom1997sparse} as the most common representative uses the absolute value or parameters as indicator for importance, but several other approaches have been proposed \cite{kusupati2020thresholdmag,bellec2017flipsign}.
Training-aware rely on data to identify parameters that have the least impact on the output, based on, e.g., first \cite{xiao2019autoprune,ding2019centripetal,lis2019full,lee2018snip,srinivas2015gate} or second order \cite{hassibi1993second,lecun1990second,dong2017learning,wang2019eigendamage,theis2018fisher,ba2016distributed,martens2015optimizing} approximations of the loss functions.
Variational methods like Variational Dropout (VD) \cite{molchanov2017variational} explicitly model the distribution of parameters and remove those with high amount of noise.

A large comparative study of sparsification methods found that simple MP can match or outperform more complicated VD on large models \cite{gale2019state}. Similarly, \cite{neuralmagic} offers a selection of sparsified large-scale NNs. 
Despite the great diversity of sparsification methods and the application of those methods to a diverse range of NNs, sparsification has not yet been applied and studied on a large population of CNNs.\looseness-1

\begin{figure*}[h!]
\centering
{\includegraphics[width = 1.9\columnwidth]{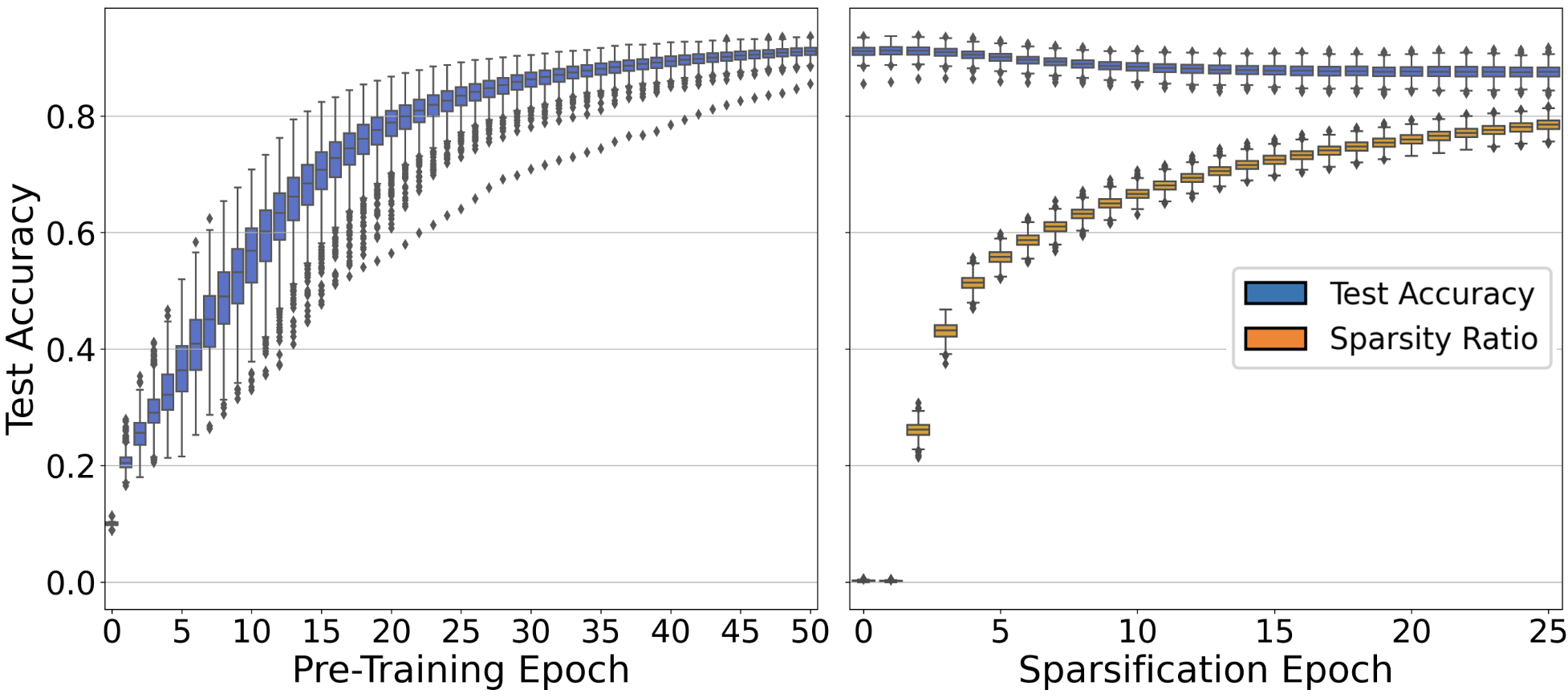}}
\caption[Sparsification frequencies]{(Left:) Test accuracy over the initial training over a fixed number of epochs in the original MNIST Seed model zoo. (Right:) Test accuracy and sparsity over the 25 epochs of VD sparsification starting from the last epoch of the original training.}
\label{fig:box_mnist_seed}
\end{figure*}
\paragraph{Populations of Neural Networks}
Recently, populations of models have become an object of study. 
Several approaches predict model properties from model features \citep{yakTaskArchitectureIndependentGeneralization2019, jiangPredictingGeneralizationGap2019,corneanuComputingTestingError2020, martinTraditionalHeavyTailedSelf2019,unterthinerPredictingNeuralNetwork2020,eilertsenClassifyingClassifierDissecting2020} or compare models based on their activations \citep{raghuSVCCASingularVector2017,morcosInsightsRepresentationalSimilarity2018,nguyenWideDeepNetworks2020}.
Other methods leverage zoos for transfer or meta learning \citep{liuKnowledgeFlowImprove2019,shuZooTuningAdaptiveTransfer2021,rameshModelZooGrowing2022}.\looseness-1

Another line of work investigates the weight space of trained models \citep{lucasAnalyzingMonotonicLinear,wortsmanLearningNeuralNetwork2021,bentonLossSurfaceSimplexes2021,ainsworthGitReBasinMerging2022,ilharco2022editing}. Recently, several methods have been proposed to learn representations of trained models \citep{denilPredictingParametersDeep2013,berardiLearningSpaceDeep2022,peeblesLearningLearnGenerative2022,ashkenazi2022nern,wang2023compact,navon2023equivariant}. \citep{schuerholt2021ssl,schurholt2022gener} proposed a self-supervised approach to learn representations of populations of models, which they dub hyper-representations and show to disentangle model properties and be useful to generate new models. 
Nonetheless, there are only few structured datasets of model zoos. \citep{gavrikovCNNFilterDB2022} publish and analyse a dataset of convolutional filters. \cite{schurholt2022model} provide a large dataset of diverse, pre-trained models, which form the basis for our sparsification work.

%
\section{Generating Sparsified Model Zoo Twins}
To analyse sparsity on populations, we apply two sparsification methods on existing pre-trained model zoos, as outlined in Figures \ref{fig:pipeline} and \ref{fig:box_mnist_seed}.
We select magnitude pruning and variational dropout as representative for data-free and training-aware methods, since they can be applied to small or medium sized CNNs that were already trained to convergence and the methods are appropriate for scaling to large populations of models. \looseness-1

The model zoos of \cite{schurholt2022model} serve as a starting point of the sparsification process. We refer to these model zoos as original model zoos. \cite{schurholt2022model} establish a setting of varying architectures $\mathcal{A}$ and hyperparameters $\mathcal{\lambda}$ on different datasets $\mathcal{D}$ for the generation of their zoos, which we adopt for this work.  The zoos were trained on \texttt{MNIST}~\citep{lecunGradientbasedLearningApplied1998}, \texttt{Fashion-MNIST}~\citep{xiaoFashionMNISTNovelImage2017}, \texttt{SVHN}~\citep{netzerReadingDigitsNatural2011}, \texttt{USPS}~\citep{hullDatabaseHandwrittenText1994}, \texttt{CIFAR-10}~\citep{krizhevskyLearningMultipleLayers2009} and \texttt{STL-10}~\citep{coatesAnalysisSingleLayerNetworks2011} using a small CNN architecture. To sparsify the model zoos, we apply both sparsification methods to the last state of each model in the zoos. To ensure that the sparsified versions of the CNNs can be compared with their original versions, the generating factors $\mathcal{A}$, $\mathcal{\lambda}$ and $\mathcal{D}$ of the original models remain unchanged, except for the learning rate. \looseness-1

\paragraph{Magnitude Pruning} 
To sparsify model zoos with MP, we select several sparsity ratios and sparsify each model in the zoo accordingly. The corresponding fraction of weights with smallest absolute value is set to zero and removed from the set of learnable parameters. We use global unstructured MP and rely on the pytorch implementation \cite{paganini}. MP generally hurts the performance, so we fine-tune the pruned models on their original dataset to recover for a fixed number of epochs. During fine-tuning we document each epoch by saving the current state dict of the model and report the test accuracy and generalization gap.\looseness-1


\paragraph{Variational Dropout} Following a similar setup, we apply VD for defined number of epochs on the last state of every model in the model zoos. Following \cite{gale2019state}, we reduce the learning rate compared to the original zoos. After training, the parameters with high variance ($\alpha \geq 3$) are removed from the NNs. As VD includes training, we do not fine-tuning the models further. We document each training epoch by saving the state dict as well es the accuracy ratio, test accuracy and generalization gap.\looseness-1

\section{Experiments}

This section outlines the experimental setup, evaluation, and analysis of generated sparsified populations of NN models.

\subsection{Experimental Setup}
We sparsify 14 model zoos with VD and 10 model zoos with MP using the the methods introduced above. In the case of MP, we sparsify each zoo with sparsity levels $[10,20,30,40,50,60,70,80,90]$\%. This is followed by 15 epochs of fine-tuning, in which the pruned weights do not receive a weight update. For our experiments, we use the pruning library of PyTorch \cite{paganini}. For VD, each weight parameter of the model receives an additional parameter $\sigma$. Each model is trained for 25 epochs and the learnable parameters $\mathbf{w}$ and $\mathbf{\sigma}$ are optimized. Both $\mathbf{w}$ and $\mathbf{\sigma}$ are aggregated in a per-parameter value $\alpha$. Weights are pruned for $\alpha > 3$. For the implementation of  VD, we adapted the fully-connected and convolutional layers of PyTorch based on the code of several previous works \cite{ryzhikov2021ard,gale2019state,molchanov2017variational}.

\paragraph{Computing Infrastructure:} The model zoos were sparsified on nodes with up to 4 CPUs and 64g RAM. Sparsifying a zoo of 1000 models takes 2-3 days. Large and more complex model zoos consisting of roughly 2600 models and greater diversity in terms of hyperparameters may take up to 11 days. Hyper-representations are trained on a GPU of a NVIDIA DGX2 station for up to 12 hours.\looseness-1

\subsection{Evaluation}
For every model at every state, we record test accuracy, generalization gap and sparsity ratio as fundamental meterics to evaluate models. Further, we compute the agreement between original and sparisified models and learn hyper-representations, to evaluate the structure of populations of sparsified models.

\paragraph{Model Agreement}
As one measure for evaluation, we compute the pairwise agreement of models within the sparsified and original model zoos. The models agree when both predict the same class given same test data. Per model pair ($k$ and $l$) this is summed up as follows:
\begin{equation} \label{eq:aggr1}
\kappa_{aggr} = \frac{1}{N} \sum_{i=1}^{N} \lambda_{y_{i}},
\end{equation}
for test samples $i = 1,...,N$, where $\lambda_{y_{i}}$ = 1, if $y_{i}^{k}$ = $y_{i}^{l}$ and $\lambda_{y_{i}}$ = 0 otherwise.

\paragraph{Hyper-Representation Learning}
For a deeper understanding of the weight spaces of the model zoos created with VD, we train a attention based auto-encoder (AE) proposed by \cite{schurholt2022gener,schurholt2022hyper}.
We learn task-agnostic hyper-representations in a self-supervised learning setting. Such representations can provide a proxy to how structured the sparsification process is. Explicitly, it provides insights in how well weights and alphas can be compressed and how well the latent space disentangles model properties like accuracy or sparsity.
We adapt the AE to take non-sparsified weights as input and reconstruct to weights and sparsification maps ($\alpha$). To improve the reconstruction quality, we introduce a new loss normalisation for the reconstruction of the alpha parameters defined as 
\begin{equation} \label{eq:recona}
\mathcal{L}_{MSE}^{\alpha} = \frac{1}{M}\sum_{i=1}^M \Big | \Big| \tanh\bigl({\frac{\mathbf{\hat{\alpha}}_i-t}{r}}\bigr) - \tanh\bigl({\frac{\mathbf{\alpha}_i-t}{r}\bigr) \Big| \Big|_2^2} ,
\end{equation}
where $t$ refers to the pruning threshold and $r$ to the selected range of interest. With that, we force the model to pay attention to the active range around the threshold that determines sparsification. Details of the model are shown in Appendix \ref{AppAE} and \ref{AppSSL}.\looseness-1

\subsection{Experimental Results and Analysis}
In this section we analyze the 24 sparsified model zoos. Due to the large scope of the results we only show highlights here and provide full details in Appendix \ref{AppResMP} and \ref{AppResVD}.\looseness-1

\paragraph{Robust Performance on Population Level:} 
As previous work investigated sparsification on single models, or hyperparameter optimization of sparsification, the robustness of sparsification methods on populations has not yet been evaluated.
Related work indicates that pruning the excess parameters of a model reduces overfitting and thus improves test accuracy and generalization \cite{hoefler2021sparsity,bartoldson2019pruningnoise}. With further increasing sparsity, functional parts of the models are removed and the performance drops. 
To investigate the performance of the methods we consider the sparsity-ratio, test accuracy and generalization gap (train accuracy - test accuracy) as metrics. 
In our experiments, magnitude pruning and variational dropout have showed remarkably robust sparsification performance on a population basis, preserving the original accuracy for considerable levels of sparsity. As illustrated in \ref{fig:box_mnist_seed}, the distribution of the performance metrics of the individual models in the zoo is very consistent and the variation from the top to the worst performing models is low. Although the standard deviation of the performance is higher on model zoos trained on a more sophisticated image dataset (e.g. CIFAR-10), comparable results are achieved. The results furthermore confirm on a population level, that the generalization gap of is lower for models with moderate sparsification levels.

\begin{figure}[H]
\vspace{10pt}
\centering
{\includegraphics[width = 1.0\columnwidth, trim= 0 50 0 50]{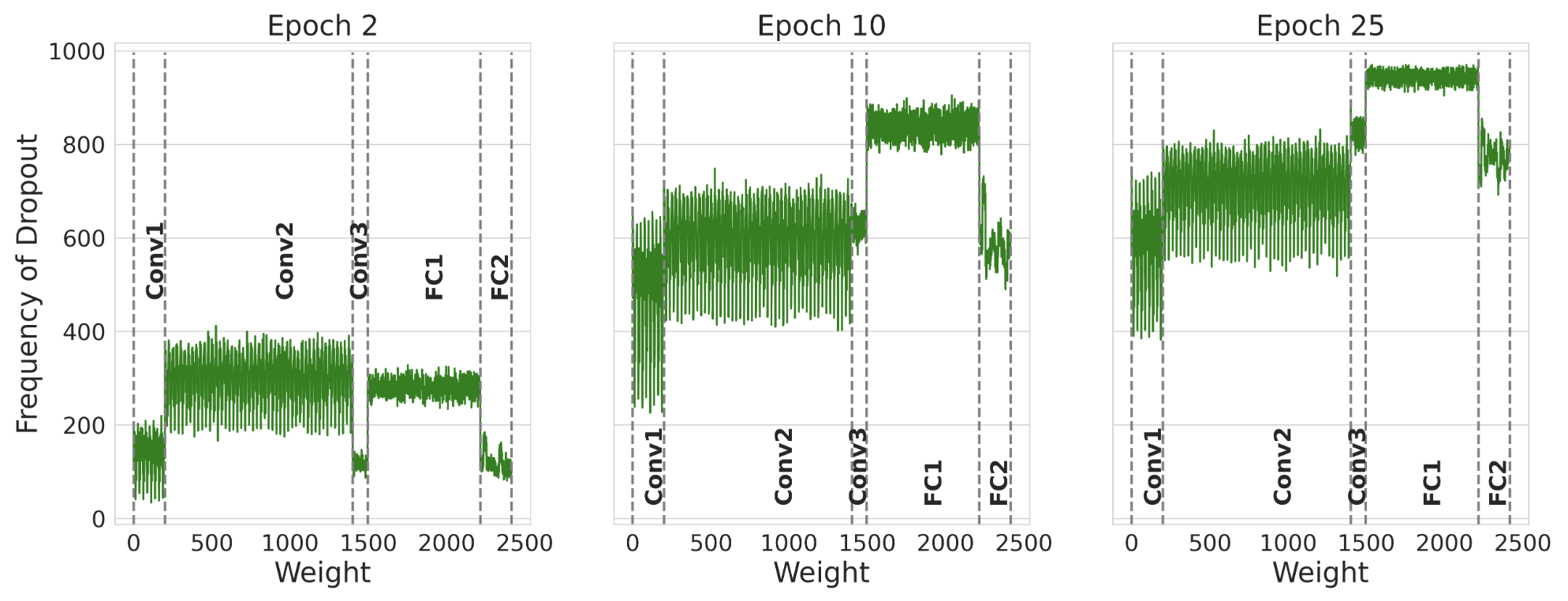}}
\caption[Sparsity per Layer]{Sparsification Frequency per weight for the MNIST zoo at different VD epochs. Within layers, there is remarkable consistency. Further, different layers are pruned in different phases}
\label{fig:freqdrop}
\end{figure}

\paragraph{Larger Layers Achieve Higher Sparsity Ratios}
The previous results indicate considerable robustness and consistency in the sparsification results within and between model zoos. To shed further light on sparsification patterns, we investigate the sparsification per layer. Within zoos, the sparsification ratios per layer are remarkably consistent, see Figure \ref{fig:freqdrop}. Across all zoos, our experiments show that larger layers are more strongly pruned, since a positive relationship between the number of parameters of a layer and the corresponding sparsity ratio exists. This relationship is shown in Figure \ref{fig:binned_scatter}. Detailed results regarding the sparsity per layer can be found in Appendix \ref{fig:binned_scatter}, \ref{AppSparLayerVD} and \ref{AppSparLayerMP}. 
This may indicate that the allocation of parameters in the architecture for the original model zoos of \cite{schurholt2022model} was not optimal. This is in line with the literature \cite{hoefler2021sparsity}, which states that pruning works particularly well for over-parameterized models.

\begin{figure}[H]
    \centering
    {\includegraphics[width = 0.8\columnwidth]{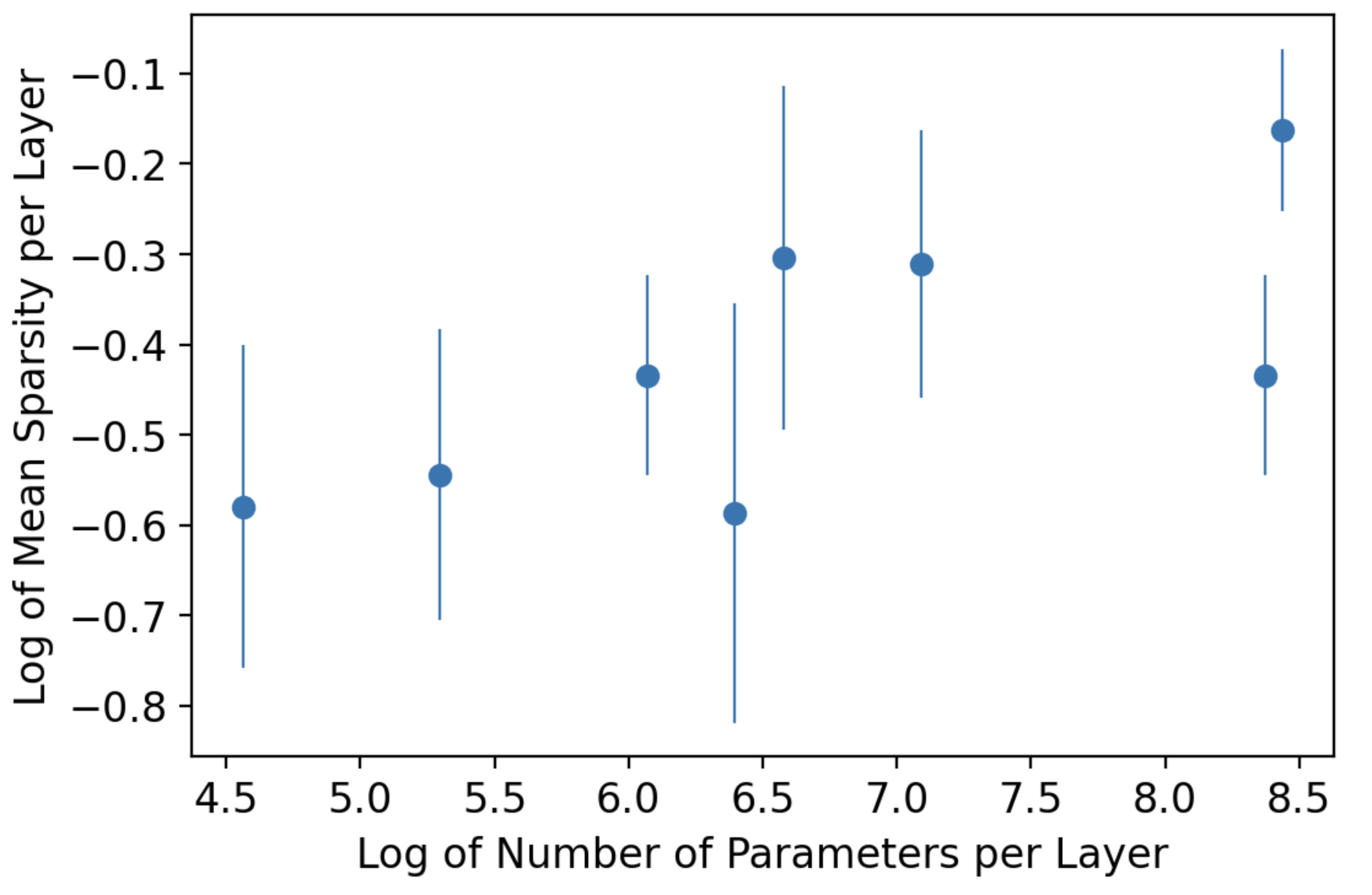}}
    \caption[Sparsity per Layer]{Binned scatter plot of all model zoos sparsified with variational dropout at epoch 5, 10, 15 and 20. Epoch 25 is not shown because certain model zoos collapsed at high sparsity ratios and this would distort the plot. The x-axis shows the logarithmized layer size, the y-axis the logarithmized mean sparsity level. The error bar represents the standard deviation of the sparsity.}
    \label{fig:binned_scatter}
\end{figure}

\begin{figure}[h!] 
\centering
{\includegraphics[width = \columnwidth]{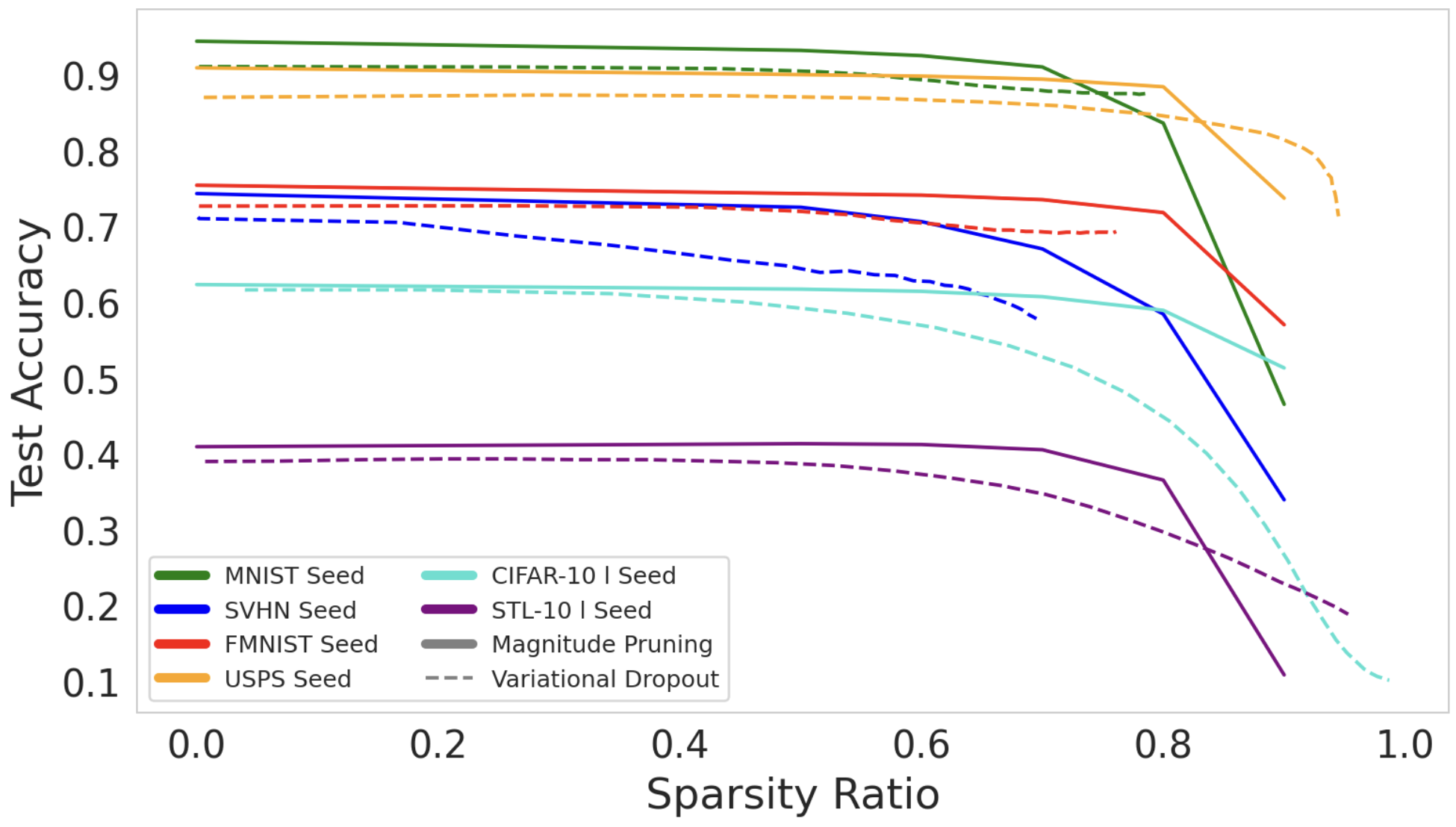}}
\caption[Mean Accuracy over Sparsity]{Mean accuracy per zoo over sparsity for a selection of model zoos sparsified with VD and MP. MP outperforms VD up to sparsity levels of 80\%. At higher sparsity, MP performance drops, VD performance is more stable.}
\label{fig:mp_vd_compare}
\end{figure}

\paragraph{Magnitude Pruning outperforms Variational Dropout}
Related work found that MP can outperform VD, especially for moderate sparsity ratios \cite{gale2019state}.
Our results confirm that on population level. MP outperforms VD for sparsification levels of up to 80\% consistently, see Figure \ref{fig:mp_vd_compare}, Appendix \ref{AppResMP} and \ref{AppResVD}.
At higher sparsification levels, MP shows steep drops in performance. VD on some zoos is more stable and thus shows higher performance at higher sparsification levels, justifying the larger parameter count and computational load.\looseness-1

\begin{table}[h!]
\caption{Agreement overview between original and twin Seed model zoos. The values mean (std) are reported in \%. Higher values indicate higher agreement. \textit{Agreem} denotes Agreement.
}
\label{table:agreement_paper}
\vspace{2mm}
\tiny
\setlength{\tabcolsep}{4pt}
\begin{tabularx}{\columnwidth}{l|ccc|ccc}
\toprule
           & \multicolumn{3}{c}{\textbf{Magnitude Pruning}} & \multicolumn{3}{c}{\textbf{Variational Dropout}} \\
\textbf{Model Zoo} & \textbf{Accuracy} & \textbf{Sparsity} & \textbf{Agreem}                  & \textbf{Accuracy} & \textbf{Sparsity}    & \textbf{Agreem}      \\
\cmidrule(lr){1-1} \cmidrule(lr){2-4} \cmidrule(lr){5-7} 
MNIST (s)      & 83.7 (13.5)         & 80.0 (0.0)            & 82.1 (13.0)             & 87.6 (1.2)      & 78.0 (1.1)           & {\textbf{83.4 (1.4)}} \\
USPS (s)       & 73.8 (17.3)          & 90.0 (0.0)            & 74.3 (17.3)             & 82.3 (1.5)      & 88.5 (0.6)           & {\textbf{86.6 (1.1)}} \\
SVHN (s)       & 70.7 (7.7)           & 60.0 (0.0)            & {\textbf{74.9 (2.5)}}   & 62.2 (7.2)      & 62.8 (2.9)           & 57.5 (6.0)           \\
FMNIST (s)     & 73.6 (1.3)           & 70.0 (0.0)            & {\textbf{79.7 (1.9)}}   & 69.3 (1.2)      & 72.0 (1.0)           & 76.2 (2.1)           \\
CIFAR-10 (s)   & 47.3 (1.2)           & 70.0 (0.0)            & {\textbf{78.4  (1.2)}}  & 40.5 (1.5)      & 67.1 (2.8)           & 61.0 (2.7)           \\
STL-10 (s)     & 40.6 (0.8)           & 70.0 (0.0)            & \textbf{55.9 (2.8)}     & 35.9 (1.2)      & 66.5 (1.3)           & 54.4 (2.8)    \\
\bottomrule
\end{tabularx}
\end{table}

\paragraph{Agreement between Twin and Original Model Zoos}
By analysing the agreement between the original and twin models, we investigate how well the two methods preserve the behavior of the original models, beyond loss or accuracy. 
The agreement is evaluated for six model zoos at the sparsity ratio 60\%, 70\%, 80\% or 90\%. The sparsity ratio was selected such that a favourable accuracy-sparsity trade-off is achieved in variational dropout.\looseness-1

The results show relatively high levels of agreement between 60 and 80 \% for both methods. Unsurprisingly, the agreement is higher for overall higher levels of accuracy. Generally, MP achieves higher accuracy and agreement, and appears to therefore preserve the original behavior of the model better. Our results indicate that simple performance metrics like accuracy may be a good proxy to estimate preserved behavior like agreement.


\paragraph{Performance of Original and Sparsified Models are Correlated}
The sparsification of populations show remarkably robust results, as indicated above. Nonetheless, there is a spread in the performance of sparsified models, see Figure \ref{fig:box_mnist_seed}.  In practice, it is relevant to identify candidates for high performance at high sparsity before sparsification. 
As first approximation, we compute the correlation between model performance before and after sparsification. We use Pearson's r as well as Kendall's tau coefficients, the former measures the covariance normalized by the product of variances, the latter measures agreement in rank order between the two paired samples.
The results show a remarkable high correlation between original and sparsified models, see Table \ref{table:performance_correlation}. For fixed sparsity levels with MP, the Pearson's r correlation is above 90\% with a single exception. The Kendall's tau is similarly high, indicating that the rank order of samples remains preserved to a high degree.
Since the sparsification levels of VD zoos are not as consistent, the correlation values are lower, but confirm the finding.
Consequently, based on the results of the sparsified populations, the best performing models will likely be the best or among the best sparsified models.
\begin{table}[h!]
\centering
\caption{Correlation between the per-model performance of original and sparsified accuracy. Values are Pearson correlation and Kendall's tau in \%. Original model performance and sparsified performance are highly correlated.  
}
\label{table:performance_correlation}
\vspace{2mm}
\tiny
\setlength{\tabcolsep}{3pt}
\begin{tabularx}{\columnwidth}{l|ccc|ccc}
\toprule
           & \multicolumn{3}{c}{\textbf{Magnitude Pruning}} & \multicolumn{3}{c}{\textbf{Variational Dropout}} \\
          \textbf{Model Zoo} & \textbf{Sparsity}   & \textbf{Pearson's r}  & \textbf{Kendall's tau}  & \textbf{Sparsity}   & \textbf{Pearson's r}   & \textbf{Kendall's tau}  \\
\cmidrule(lr){1-1} \cmidrule(lr){2-4} \cmidrule(lr){5-7} 
MNIST     & 80.0       & 99.7     & 89.4           & 78.0       & 51.4      & 35.2           \\
USPS      & 80.0       & 90.0     & 71.6           & 88.5       & 52.1      & 35.0           \\
SVHN      & 80.0       & 98.7     & 92.0           & 64.7       & 91.6      & 54.7           \\
FMNIST    & 80.0       & 95.7     & 72.7           & 72.6       & 58.6      & 40.9           \\
CIFAR s   & 80.0       & 97.2     & 82.7           & 67.1       & 72.6      & 45.2           \\
CIFAR l   & 80.0       & 93.4     & 76.3           & 67.3       & 50.0      & 33.3           \\
STL s     & 80.0       & 96.7     & 82.3           & 66.5       & 56.9      & 37.7           \\
STL l & 80.0       & 73.7     & 52.8           & 49.0       & 80.9      & 61.3 \\
\bottomrule
\end{tabularx}
\end{table}

\paragraph{Disentangled Representations learned from Weight Space}
With the revised AE and its novel loss normalisation we are able to not only reconstruct the weight spaces of the CNNs but also the alpha parameters needed for the pruning decision in VD. 
The results are remarkable in that both accuracy and sparsity are highly predictable and thus disentangled very well in latent space. What is more, both weights as well as alphas are reconstructed well, indicating a high degree of structure in populations of sparsified models. This opens the door for future attempts to zero-shot sparsify models impressing such structure on pre-trained models. The results are shown in Table \ref{table:hyperreps}.


\begin{table}[h!]
\centering
\caption[Results Hyper-Representations]{Results overview of the hyper-representation learning task on the MNIST Seed and SVHN Seed model zoos. The AE is trained with the weight spaces of the original and sparsified model zoos. All values are $R^2$ reported in \%. Higher values are better. Weights and alphas are the reconstructions $R^2$. Accuracy, sparsity, epoch and generalization gap (GGap) are predicted from the embeddings as in \cite{schurholt2021self}.}
\label{table:hyperreps}
\vspace{2mm}
\tiny
\setlength{\tabcolsep}{7pt}
\begin{tabularx}{\columnwidth}{lcccccc}
\toprule
Zoo   & Weights & Alphas & Accuracy & Sparsity & Epoch & GGap \\ \midrule
MNIST (s) & 78.7    & 79.4   & 96.3     & 98.0     & 53.2  & 19.1    \\
SVHN (s) & 74.8    & 74.2   & 94.5     & 97.6     & 30.2  & 10.5    \\ \bottomrule
\end{tabularx}
\end{table}

%
\section{Conclusion}
In this work, we have analyzed sparsification on large populations of neural networks. Using magnitude pruning and variational dropout as underlying sparsification approach, we have created ten sparsified model zoo twins representing common computer vision datasets. In total, we have created 23'920 sparsified models with 1'726'000 documented model states. 
We can confirm, that both approaches - magnitude pruning (MP) and variational dropout (VD) - perform well on population level with respect to sparsification ratio and accuracy. For sparsification ratios below 80\%, MP outperforms VD. At higher sparsification ratios, both methods degrade, but VD is more stable. 
Sparsified models show high agreement with their original models, with no clear preference between the two sparsification approaches. 
We further find that performance before and after sparsification is highly correlated, indicating that the best performing model is the best candidate for sparsification.
The sparsification characteristics per layer within the zoos are surprisingly consistent. This gives rise to learning hyper-representations on sparsified model zoos, which shows to be unexpectedly successful. That indicates that sparsification is highly structured, which may be exploited for zero-shot sparsification.


\bibliography{./bibliography.bib,./bib_extra.bib}

\begin{thebibliography}{69}
\providecommand{\natexlab}[1]{#1}
\providecommand{\url}[1]{\texttt{#1}}
\expandafter\ifx\csname urlstyle\endcsname\relax
  \providecommand{\doi}[1]{doi: #1}\else
  \providecommand{\doi}{doi: \begingroup \urlstyle{rm}\Url}\fi

\bibitem[Ainsworth et~al.(2022)Ainsworth, Hayase, and
  Srinivasa]{ainsworthGitReBasinMerging2022}
Ainsworth, S.~K., Hayase, J., and Srinivasa, S.
\newblock Git {{Re-Basin}}: {{Merging Models}} modulo {{Permutation
  Symmetries}}, September 2022.

\bibitem[Ashkenazi et~al.(2022)Ashkenazi, Rimon, Vainshtein, Levi, Richardson,
  Mintz, and Treister]{ashkenazi2022nern}
Ashkenazi, M., Rimon, Z., Vainshtein, R., Levi, S., Richardson, E., Mintz, P.,
  and Treister, E.
\newblock Nern--learning neural representations for neural networks.
\newblock \emph{arXiv preprint arXiv:2212.13554}, 2022.

\bibitem[Ba et~al.(2016)Ba, Grosse, and Martens]{ba2016distributed}
Ba, J., Grosse, R., and Martens, J.
\newblock Distributed second-order optimization using kronecker-factored
  approximations.
\newblock 2016.

\bibitem[Bartoldson et~al.(2020)Bartoldson, Morcos, Barbu, and
  Erlebacher]{bartoldson2019pruningnoise}
Bartoldson, B., Morcos, A., Barbu, A., and Erlebacher, G.
\newblock The generalization-stability tradeoff in neural network pruning.
\newblock \emph{Advances in Neural Information Processing Systems},
  33:\penalty0 20852--20864, 2020.

\bibitem[Bellec et~al.(2017)Bellec, Kappel, Maass, and
  Legenstein]{bellec2017flipsign}
Bellec, G., Kappel, D., Maass, W., and Legenstein, R.
\newblock Deep rewiring: Training very sparse deep networks, 2017.
\newblock URL \url{https://arxiv.org/abs/1711.05136}.

\bibitem[Benton et~al.(2021)Benton, Maddox, Lotfi, and
  Wilson]{bentonLossSurfaceSimplexes2021}
Benton, G.~W., Maddox, W.~J., Lotfi, S., and Wilson, A.~G.
\newblock Loss {{Surface Simplexes}} for {{Mode Connecting Volumes}} and {{Fast
  Ensembling}}.
\newblock In \emph{{{PMLR}}}, 2021.

\bibitem[Berardi et~al.(2022)Berardi, De~Luigi, Salti, and
  Di~Stefano]{berardiLearningSpaceDeep2022}
Berardi, G., De~Luigi, L., Salti, S., and Di~Stefano, L.
\newblock Learning the {{Space}} of {{Deep Models}}, June 2022.

\bibitem[Blalock et~al.(2020)Blalock, Gonzalez~Ortiz, Frankle, and
  Guttag]{blalock2020state}
Blalock, D., Gonzalez~Ortiz, J.~J., Frankle, J., and Guttag, J.
\newblock What is the state of neural network pruning?
\newblock \emph{Proceedings of machine learning and systems}, 2:\penalty0
  129--146, 2020.

\bibitem[Brock et~al.(2018)Brock, Donahue, and Simonyan]{brock2018scale}
Brock, A., Donahue, J., and Simonyan, K.
\newblock Large scale gan training for high fidelity natural image synthesis.
\newblock \emph{arXiv preprint arXiv:1809.11096}, 2018.

\bibitem[Chen et~al.(2020)Chen, Kornblith, Norouzi, and Hinton]{chen2020ntxent}
Chen, T., Kornblith, S., Norouzi, M., and Hinton, G.
\newblock A simple framework for contrastive learning of visual
  representations, 2020.
\newblock URL \url{https://arxiv.org/abs/2002.05709}.

\bibitem[Coates et~al.(2011)Coates, Lee, and
  Ng]{coatesAnalysisSingleLayerNetworks2011}
Coates, A., Lee, H., and Ng, A.~Y.
\newblock An {{Analysis}} of {{Single-Layer Networks}} in {{Unsupervised
  Feature Learning}}.
\newblock In \emph{Proceedings of the 14th {{International Con-}} Ference on
  {{Artificial Intelligence}} and {{Statistics}} ({{AISTATS}})}, pp.\ ~9, 2011.

\bibitem[Corneanu et~al.(2020)Corneanu, Escalera, and
  Martinez]{corneanuComputingTestingError2020}
Corneanu, C.~A., Escalera, S., and Martinez, A.~M.
\newblock Computing the {{Testing Error Without}} a {{Testing Set}}.
\newblock In \emph{2020 {{IEEE}}/{{CVF Conference}} on {{Computer Vision}} and
  {{Pattern Recognition}} ({{CVPR}})}, pp.\  2674--2682, {Seattle, WA, USA},
  June 2020. {IEEE}.
\newblock ISBN 978-1-72817-168-5.
\newblock \doi{10.1109/CVPR42600.2020.00275}.

\bibitem[Cun et~al.(1990)Cun, Denker, and Solla]{lecun1990second}
Cun, Y.~L., Denker, J.~S., and Solla, S.~A.
\newblock \emph{Optimal Brain Damage}, pp.\  598–605.
\newblock Morgan Kaufmann Publishers Inc., San Francisco, CA, USA, 1990.
\newblock ISBN 1558601007.

\bibitem[Denil et~al.(2013)Denil, Shakibi, Dinh, and
  Ranzato]{denilPredictingParametersDeep2013}
Denil, M., Shakibi, B., Dinh, L., and Ranzato, M.
\newblock Predicting {{Parameters}} in {{Deep Learning}}.
\newblock In \emph{Neural {{Information Processing Systems}} ({{NeurIPS}})},
  pp.\ ~9, 2013.

\bibitem[Ding et~al.(2019)Ding, Ding, Guo, and Han]{ding2019centripetal}
Ding, X., Ding, G., Guo, Y., and Han, J.
\newblock Centripetal sgd for pruning very deep convolutional networks with
  complicated structure.
\newblock In \emph{Proceedings of the IEEE/CVF conference on computer vision
  and pattern recognition}, pp.\  4943--4953, 2019.

\bibitem[Dong et~al.(2017)Dong, Chen, and Pan]{dong2017learning}
Dong, X., Chen, S., and Pan, S.
\newblock Learning to prune deep neural networks via layer-wise optimal brain
  surgeon.
\newblock \emph{Advances in Neural Information Processing Systems}, 30, 2017.

\bibitem[Eilertsen et~al.(2020)Eilertsen, J{\"o}nsson, Ropinski, Unger, and
  Ynnerman]{eilertsenClassifyingClassifierDissecting2020}
Eilertsen, G., J{\"o}nsson, D., Ropinski, T., Unger, J., and Ynnerman, A.
\newblock Classifying the classifier: Dissecting the weight space of neural
  networks.
\newblock \emph{arXiv:2002.05688 [cs]}, February 2020.

\bibitem[Frantar \& Alistarh(2023)Frantar and Alistarh]{sparsegpt2023frantar}
Frantar, E. and Alistarh, D.
\newblock Sparsegpt: Massive language models can be accurately pruned in
  one-shot, 2023.
\newblock URL \url{https://arxiv.org/abs/2301.00774}.

\bibitem[Gale et~al.(2019)Gale, Elsen, and Hooker]{gale2019state}
Gale, T., Elsen, E., and Hooker, S.
\newblock The state of sparsity in deep neural networks.
\newblock \emph{arXiv preprint arXiv:1902.09574}, 2019.

\bibitem[Gavrikov \& Keuper(2022)Gavrikov and Keuper]{gavrikovCNNFilterDB2022}
Gavrikov, P. and Keuper, J.
\newblock {{CNN Filter DB}}: {{An Empirical Investigation}} of {{Trained
  Convolutional Filters}}.
\newblock In \emph{Proceedings of the {{IEEE}}/{{CVF Conference}} on {{Computer
  Vision}} and {{Pattern Recognition}} ({{CVPR}})}, pp.\ ~11, 2022.

\bibitem[Giuffrida et~al.(2021)Giuffrida, Fanucci, Meoni, Bati{\v{c}}, Buckley,
  Dunne, van Dijk, Esposito, Hefele, Vercruyssen, et~al.]{giuffrida2021varphi}
Giuffrida, G., Fanucci, L., Meoni, G., Bati{\v{c}}, M., Buckley, L., Dunne, A.,
  van Dijk, C., Esposito, M., Hefele, J., Vercruyssen, N., et~al.
\newblock The $\phi$-sat-1 mission: the first on-board deep neural network
  demonstrator for satellite earth observation.
\newblock \emph{IEEE Transactions on Geoscience and Remote Sensing},
  60:\penalty0 1--14, 2021.

\bibitem[Han et~al.(2015)Han, Mao, and Dally]{han2015deep}
Han, S., Mao, H., and Dally, W.~J.
\newblock Deep compression: Compressing deep neural networks with pruning,
  trained quantization and huffman coding.
\newblock \emph{arXiv preprint arXiv:1510.00149}, 2015.

\bibitem[Hassibi et~al.(1993)Hassibi, Stork, Wolff, and
  Watanabe]{hassibi1993second}
Hassibi, B., Stork, D.~G., Wolff, G., and Watanabe, T.
\newblock Optimal brain surgeon: Extensions and performance comparisons.
\newblock In \emph{Proceedings of the 6th International Conference on Neural
  Information Processing Systems}, NIPS'93, pp.\  263–270, San Francisco, CA,
  USA, 1993. Morgan Kaufmann Publishers Inc.

\bibitem[Hoefler et~al.(2021)Hoefler, Alistarh, Ben-Nun, Dryden, and
  Peste]{hoefler2021sparsity}
Hoefler, T., Alistarh, D., Ben-Nun, T., Dryden, N., and Peste, A.
\newblock Sparsity in deep learning: Pruning and growth for efficient inference
  and training in neural networks.
\newblock \emph{J. Mach. Learn. Res.}, 22\penalty0 (241):\penalty0 1--124,
  2021.

\bibitem[Howard et~al.(2017)Howard, Zhu, Chen, Kalenichenko, Wang, Weyand,
  Andreetto, and Adam]{howard2017mobilenets}
Howard, A.~G., Zhu, M., Chen, B., Kalenichenko, D., Wang, W., Weyand, T.,
  Andreetto, M., and Adam, H.
\newblock Mobilenets: Efficient convolutional neural networks for mobile vision
  applications.
\newblock \emph{arXiv preprint arXiv:1704.04861}, 2017.

\bibitem[Hull(1994)]{hullDatabaseHandwrittenText1994}
Hull, J.
\newblock A database for handwritten text recognition research.
\newblock \emph{IEEE Transactions on Pattern Analysis and Machine
  Intelligence}, 16\penalty0 (5):\penalty0 550--554, May 1994.
\newblock ISSN 1939-3539.
\newblock \doi{10.1109/34.291440}.

\bibitem[Ilharco et~al.(2022)Ilharco, Ribeiro, Wortsman, Gururangan, Schmidt,
  Hajishirzi, and Farhadi]{ilharco2022editing}
Ilharco, G., Ribeiro, M.~T., Wortsman, M., Gururangan, S., Schmidt, L.,
  Hajishirzi, H., and Farhadi, A.
\newblock Editing models with task arithmetic.
\newblock \emph{arXiv preprint arXiv:2212.04089}, 2022.

\bibitem[Jiang et~al.(2019)Jiang, Krishnan, Mobahi, and
  Bengio]{jiangPredictingGeneralizationGap2019}
Jiang, Y., Krishnan, D., Mobahi, H., and Bengio, S.
\newblock Predicting the {{Generalization Gap}} in {{Deep Networks}} with
  {{Margin Distributions}}.
\newblock \emph{arXiv:1810.00113 [cs, stat]}, June 2019.

\bibitem[Kaplan et~al.(2020)Kaplan, McCandlish, Henighan, Brown, Chess, Child,
  Gray, Radford, Wu, and Amodei]{kaplan2020scaling}
Kaplan, J., McCandlish, S., Henighan, T., Brown, T.~B., Chess, B., Child, R.,
  Gray, S., Radford, A., Wu, J., and Amodei, D.
\newblock Scaling laws for neural language models.
\newblock \emph{arXiv preprint arXiv:2001.08361}, 2020.

\bibitem[Krizhevsky(2009)]{krizhevskyLearningMultipleLayers2009}
Krizhevsky, A.
\newblock Learning {{Multiple Layers}} of {{Features}} from {{Tiny Images}}.
\newblock pp.\ ~60, 2009.

\bibitem[Kusupati et~al.(2020)Kusupati, Ramanujan, Somani, Wortsman, Jain,
  Kakade, and Farhadi]{kusupati2020thresholdmag}
Kusupati, A., Ramanujan, V., Somani, R., Wortsman, M., Jain, P., Kakade, S.,
  and Farhadi, A.
\newblock Soft threshold weight reparameterization for learnable sparsity.
\newblock In \emph{International Conference on Machine Learning}, pp.\
  5544--5555. PMLR, 2020.

\bibitem[LeCun et~al.(1998)LeCun, Bottou, Bengio, and
  Haffner]{lecunGradientbasedLearningApplied1998}
LeCun, Y., Bottou, L., Bengio, Y., and Haffner, P.
\newblock Gradient-{{Based Learning Applied}} to {{Document Recognition}}.
\newblock \emph{Proceedings of the IEEE}, 86\penalty0 (11):\penalty0
  2278--2324, November 1998.

\bibitem[Lee et~al.(2018)Lee, Ajanthan, and Torr]{lee2018snip}
Lee, N., Ajanthan, T., and Torr, P. H.~S.
\newblock Snip: Single-shot network pruning based on connection sensitivity,
  2018.
\newblock URL \url{https://arxiv.org/abs/1810.02340}.

\bibitem[Lis et~al.(2019)Lis, Golub, and Lemieux]{lis2019full}
Lis, M., Golub, M., and Lemieux, G.
\newblock Full deep neural network training on a pruned weight budget.
\newblock \emph{Proceedings of Machine Learning and Systems}, 1:\penalty0
  252--263, 2019.

\bibitem[Liu et~al.(2019)Liu, Peng, and Schwing]{liuKnowledgeFlowImprove2019}
Liu, I.-J., Peng, J., and Schwing, A.~G.
\newblock Knowledge {{Flow}}: {{Improve Upon Your Teachers}}.
\newblock In \emph{International {{Conference}} on {{Learning Representations}}
  ({{ICLR}})}, April 2019.

\bibitem[Lucas et~al.()Lucas, Bae, Zhang, Fort, Zemel, and
  Grosse]{lucasAnalyzingMonotonicLinear}
Lucas, J., Bae, J., Zhang, M.~R., Fort, S., Zemel, R., and Grosse, R.
\newblock Analyzing {{Monotonic Linear Interpolation}} in {{Neural Network Loss
  Landscapes}}.
\newblock pp.\ ~12.

\bibitem[Martens \& Grosse(2015)Martens and Grosse]{martens2015optimizing}
Martens, J. and Grosse, R.
\newblock Optimizing neural networks with kronecker-factored approximate
  curvature.
\newblock In \emph{International conference on machine learning}, pp.\
  2408--2417. PMLR, 2015.

\bibitem[Martin \& Mahoney(2019)Martin and
  Mahoney]{martinTraditionalHeavyTailedSelf2019}
Martin, C.~H. and Mahoney, M.~W.
\newblock Traditional and {{Heavy-Tailed Self Regularization}} in {{Neural
  Network Models}}.
\newblock \emph{arXiv:1901.08276 [cs, stat]}, January 2019.

\bibitem[Molchanov et~al.(2017)Molchanov, Ashukha, and
  Vetrov]{molchanov2017variational}
Molchanov, D., Ashukha, A., and Vetrov, D.
\newblock Variational dropout sparsifies deep neural networks.
\newblock In \emph{International Conference on Machine Learning}, pp.\
  2498--2507. PMLR, 2017.

\bibitem[Morcos et~al.(2018)Morcos, Raghu, and
  Bengio]{morcosInsightsRepresentationalSimilarity2018}
Morcos, A.~S., Raghu, M., and Bengio, S.
\newblock Insights on representational similarity in neural networks with
  canonical correlation.
\newblock \emph{arXiv:1806.05759 [cs, stat]}, June 2018.

\bibitem[Navon et~al.(2023)Navon, Shamsian, Achituve, Fetaya, Chechik, and
  Maron]{navon2023equivariant}
Navon, A., Shamsian, A., Achituve, I., Fetaya, E., Chechik, G., and Maron, H.
\newblock Equivariant architectures for learning in deep weight spaces.
\newblock \emph{arXiv preprint arXiv:2301.12780}, 2023.

\bibitem[Netzer et~al.(2011)Netzer, Wang, Coates, Bissacco, Wu, and
  Ng]{netzerReadingDigitsNatural2011}
Netzer, Y., Wang, T., Coates, A., Bissacco, A., Wu, B., and Ng, A.~Y.
\newblock Reading {{Digits}} in {{Natural Images}} with {{Unsupervised Feature
  Learning}}.
\newblock In \emph{{{NIPS Workshop}} on {{Deep Learning}} and {{Unsupervised
  Feature Learning}} 2011}, pp.\ ~9, 2011.

\bibitem[neuralmagic()]{neuralmagic}
neuralmagic.
\newblock Sparsezoo.
\newblock URL \url{https://sparsezoo.neuralmagic.com/}.

\bibitem[Nguyen et~al.(2020)Nguyen, Raghu, and
  Kornblith]{nguyenWideDeepNetworks2020}
Nguyen, T., Raghu, M., and Kornblith, S.
\newblock Do {{Wide}} and {{Deep Networks Learn}} the {{Same Things}}?
  {{Uncovering How Neural Network Representations Vary}} with {{Width}} and
  {{Depth}}.
\newblock \emph{arXiv:2010.15327 [cs]}, October 2020.

\bibitem[Paganini(2019)]{paganini}
Paganini, M.
\newblock Pruning tutorial¶, 2019.
\newblock URL
  \url{https://pytorch.org/tutorials/intermediate/pruning_tutorial.html}.

\bibitem[Peebles et~al.(2022)Peebles, Radosavovic, Brooks, Efros, and
  Malik]{peeblesLearningLearnGenerative2022}
Peebles, W., Radosavovic, I., Brooks, T., Efros, A.~A., and Malik, J.
\newblock Learning to {{Learn}} with {{Generative Models}} of {{Neural Network
  Checkpoints}}, September 2022.

\bibitem[Raghu et~al.(2017)Raghu, Gilmer, Yosinski, and
  {Sohl-Dickstein}]{raghuSVCCASingularVector2017}
Raghu, M., Gilmer, J., Yosinski, J., and {Sohl-Dickstein}, J.
\newblock {{SVCCA}}: {{Singular Vector Canonical Correlation Analysis}} for
  {{Deep Learning Dynamics}} and {{Interpretability}}.
\newblock \emph{arXiv:1706.05806 [cs, stat]}, June 2017.

\bibitem[Ramesh \& Chaudhari(2022)Ramesh and
  Chaudhari]{rameshModelZooGrowing2022}
Ramesh, R. and Chaudhari, P.
\newblock Model {{Zoo}}: {{A Growing}} "{{Brain}}" {{That Learns Continually}}.
\newblock In \emph{International {{Conference}} on {{Learning Representations
  ICLR}}}, 2022.

\bibitem[Ryzhikov(2021)]{ryzhikov2021ard}
Ryzhikov, A.
\newblock Pytorch ard: Pytorch implementation of variational dropout sparsifies
  deep neural networks, Nov 2021.
\newblock URL \url{https://github.com/HolyBayes/pytorch_ard}.

\bibitem[Sch{\"u}rholt et~al.(2021{\natexlab{a}})Sch{\"u}rholt, Kostadinov, and
  Borth]{schuerholt2021ssl}
Sch{\"u}rholt, K., Kostadinov, D., and Borth, D.
\newblock Self-supervised representation learning on neural network weights for
  model characteristic prediction.
\newblock \emph{Advances in Neural Information Processing Systems},
  34:\penalty0 16481--16493, 2021{\natexlab{a}}.

\bibitem[Sch{\"u}rholt et~al.(2021{\natexlab{b}})Sch{\"u}rholt, Kostadinov, and
  Borth]{schurholt2021self}
Sch{\"u}rholt, K., Kostadinov, D., and Borth, D.
\newblock Self-supervised representation learning on neural network weights for
  model characteristic prediction.
\newblock \emph{Advances in Neural Information Processing Systems},
  34:\penalty0 16481--16493, 2021{\natexlab{b}}.

\bibitem[Sch{\"u}rholt et~al.(2022{\natexlab{a}})Sch{\"u}rholt, Knyazev,
  Gir{\'o}-i Nieto, and Borth]{schurholt2022gener}
Sch{\"u}rholt, K., Knyazev, B., Gir{\'o}-i Nieto, X., and Borth, D.
\newblock Hyper-representations as generative models: Sampling unseen neural
  network weights.
\newblock \emph{Conference on Neural Information Processing Systems (NeurIPS),
  2022}, 2022{\natexlab{a}}.

\bibitem[Sch{\"u}rholt et~al.(2022{\natexlab{b}})Sch{\"u}rholt, Knyazev,
  Gir{\'o}-i Nieto, and Borth]{schurholt2022hyper}
Sch{\"u}rholt, K., Knyazev, B., Gir{\'o}-i Nieto, X., and Borth, D.
\newblock Hyper-representation for pre-training and transfer learning.
\newblock In \emph{First Workshop on Pre-training: Perspectives, Pitfalls, and
  Paths Forward at ICML 2022}, 2022{\natexlab{b}}.

\bibitem[Sch{\"u}rholt et~al.(2022{\natexlab{c}})Sch{\"u}rholt, Taskiran,
  Knyazev, Gir{\'o}-i Nieto, and Borth]{schurholt2022model}
Sch{\"u}rholt, K., Taskiran, D., Knyazev, B., Gir{\'o}-i Nieto, X., and Borth,
  D.
\newblock Model zoos: A dataset of diverse populations of neural network
  models.
\newblock \emph{Conference on Neural Information Processing Systems (NeurIPS),
  Datasets and Benchmarks Track, 2022}, 2022{\natexlab{c}}.

\bibitem[Shu et~al.(2021)Shu, Kou, Cao, Wang, and
  Long]{shuZooTuningAdaptiveTransfer2021}
Shu, Y., Kou, Z., Cao, Z., Wang, J., and Long, M.
\newblock Zoo-{{Tuning}}: {{Adaptive Transfer}} from a {{Zoo}} of {{Models}}.
\newblock In \emph{International {{Conference}} on {{Machine Learning}}
  ({{ICML}})}, pp.\ ~12, 2021.

\bibitem[Srinivas \& Babu(2015)Srinivas and Babu]{srinivas2015gate}
Srinivas, S. and Babu, R.~V.
\newblock Learning neural network architectures using backpropagation, 2015.
\newblock URL \url{https://arxiv.org/abs/1511.05497}.

\bibitem[Str{\"o}m(1997)]{strom1997sparse}
Str{\"o}m, N.
\newblock Sparse connection and pruning in large dynamic artificial neural
  networks.
\newblock In \emph{Fifth European Conference on Speech Communication and
  Technology}. Citeseer, 1997.

\bibitem[Strubell et~al.(2019)Strubell, Ganesh, and
  McCallum]{strubell2019energy}
Strubell, E., Ganesh, A., and McCallum, A.
\newblock Energy and policy considerations for deep learning in nlp, 2019.
\newblock URL \url{https://arxiv.org/abs/1906.02243}.

\bibitem[Tan \& Le(2019)Tan and Le]{tan2019size}
Tan, M. and Le, Q.
\newblock Efficientnet: Rethinking model scaling for convolutional neural
  networks.
\newblock In \emph{International conference on machine learning}, pp.\
  6105--6114. PMLR, 2019.

\bibitem[Theis et~al.(2018)Theis, Korshunova, Tejani, and
  Huszár]{theis2018fisher}
Theis, L., Korshunova, I., Tejani, A., and Huszár, F.
\newblock Faster gaze prediction with dense networks and fisher pruning, 2018.
\newblock URL \url{https://arxiv.org/abs/1801.05787}.

\bibitem[Thompson et~al.(2022)Thompson, Greenewald, Lee, and
  Manso]{thompson2022}
Thompson, N.~C., Greenewald, K., Lee, K., and Manso, G.~F.
\newblock Deep learning's diminishing returns, Nov 2022.
\newblock URL \url{https://spectrum.ieee.org/deep-learning-computational-cost}.

\bibitem[Unterthiner et~al.(2020)Unterthiner, Keysers, Gelly, Bousquet, and
  Tolstikhin]{unterthinerPredictingNeuralNetwork2020}
Unterthiner, T., Keysers, D., Gelly, S., Bousquet, O., and Tolstikhin, I.
\newblock Predicting {{Neural Network Accuracy}} from {{Weights}}.
\newblock \emph{arXiv:2002.11448 [cs, stat]}, February 2020.

\bibitem[Wang et~al.(2019)Wang, Grosse, Fidler, and Zhang]{wang2019eigendamage}
Wang, C., Grosse, R., Fidler, S., and Zhang, G.
\newblock Eigendamage: Structured pruning in the kronecker-factored eigenbasis.
\newblock In \emph{International Conference on Machine Learning}, pp.\
  6566--6575. PMLR, 2019.

\bibitem[Wang et~al.(2023)Wang, Chen, Yu, Cheung, and LeCun]{wang2023compact}
Wang, J., Chen, Y., Yu, S.~X., Cheung, B., and LeCun, Y.
\newblock Compact and optimal deep learning with recurrent parameter
  generators.
\newblock In \emph{Proceedings of the IEEE/CVF Winter Conference on
  Applications of Computer Vision}, pp.\  3900--3910, 2023.

\bibitem[Wortsman et~al.(2021)Wortsman, Horton, Guestrin, Farhadi, and
  Rastegari]{wortsmanLearningNeuralNetwork2021}
Wortsman, M., Horton, M.~C., Guestrin, C., Farhadi, A., and Rastegari, M.
\newblock Learning {{Neural Network Subspaces}}.
\newblock In \emph{International {{Conference}} on {{Machine Learning}}}, pp.\
  11217--11227. {PMLR}, July 2021.

\bibitem[Xiao et~al.(2017)Xiao, Rasul, and
  Vollgraf]{xiaoFashionMNISTNovelImage2017}
Xiao, H., Rasul, K., and Vollgraf, R.
\newblock Fashion-{{MNIST}}: A {{Novel Image Dataset}} for {{Benchmarking
  Machine Learning Algorithms}}, September 2017.

\bibitem[Xiao et~al.(2019)Xiao, Wang, and Rajasekaran]{xiao2019autoprune}
Xiao, X., Wang, Z., and Rajasekaran, S.
\newblock Autoprune: Automatic network pruning by regularizing auxiliary
  parameters.
\newblock \emph{Advances in neural information processing systems}, 32, 2019.

\bibitem[Yak et~al.(2019)Yak, Gonzalvo, and
  Mazzawi]{yakTaskArchitectureIndependentGeneralization2019}
Yak, S., Gonzalvo, J., and Mazzawi, H.
\newblock Towards {{Task}} and {{Architecture-Independent Generalization Gap
  Predictors}}.
\newblock \emph{arXiv:1906.01550 [cs, stat]}, June 2019.

\bibitem[Yu et~al.(2022)Yu, Wang, Vasudevan, Yeung, Seyedhosseini, and
  Wu]{yu2022coca}
Yu, J., Wang, Z., Vasudevan, V., Yeung, L., Seyedhosseini, M., and Wu, Y.
\newblock Coca: Contrastive captioners are image-text foundation models, 2022.
\newblock URL \url{https://arxiv.org/abs/2205.01917}.

\end{thebibliography}
\bibliographystyle{icml2023}

\newpage
\appendix
\onecolumn

\section{Experimental Overview}
\label{AppExpOv}
\begin{table}[H]
\scriptsize
\centering
\begin{tabular}{llllllll}
\toprule
\textbf{Model Zoo}  & \multicolumn{1}{c}{\textbf{W}}   & \multicolumn{1}{c}{\textbf{Activation}} & \multicolumn{1}{c}{\textbf{Optim}} & \multicolumn{1}{c}{\textbf{LR VD}}    & \multicolumn{1}{c}{\textbf{LR MP}}    & \multicolumn{1}{c}{\textbf{WD}} & \multicolumn{1}{c}{\textbf{Dropout}} \\
\midrule
MNIST (s) Seed      & \multicolumn{1}{c}{{2416}}  & \multicolumn{1}{c}{T}                   & \multicolumn{1}{c}{AD}             & \multicolumn{1}{c}{3e-4}          & \multicolumn{1}{c}{\textbf{1e-3}} & \multicolumn{1}{c}{0}           & \multicolumn{1}{c}{0}                \\
MNIST (s) Random    & \multicolumn{1}{c}{{2416}}  & \multicolumn{1}{c}{T, S, R, G}          & \multicolumn{1}{c}{AD, SGD}        & \multicolumn{1}{c}{1e-3, 1e-4}        & \multicolumn{1}{c}{1e-3, 1e-4}        & \multicolumn{1}{c}{1e-3, 1e-4}  & \multicolumn{1}{c}{0, 0.5}           \\
MNIST (s) Fixed     & \multicolumn{1}{c}{{2416}}  & \multicolumn{1}{c}{T, S, R, G}          & \multicolumn{1}{c}{AD, SGD}        & \multicolumn{1}{c}{1e-3, 1e-4}        & \multicolumn{1}{c}{1e-3, 1e-4}        & \multicolumn{1}{c}{1e-3, 1e-4}  & \multicolumn{1}{c}{0, 0.5}           \\
SVHN (s) Seed       & \multicolumn{1}{c}{{2416}}  & \multicolumn{1}{c}{T}                   & \multicolumn{1}{c}{AD}             & \multicolumn{1}{c}{3e-3}          & \multicolumn{1}{c}{\textbf{1e-3}} & \multicolumn{1}{c}{0}           & \multicolumn{1}{c}{0}                \\
FMNIST (s) Seed     & \multicolumn{1}{c}{{2416}}  & \multicolumn{1}{c}{T}                   & \multicolumn{1}{c}{AD}             & \multicolumn{1}{c}{3e4}          & \multicolumn{1}{c}{\textbf{1e-3}} & \multicolumn{1}{c}{0}           & \multicolumn{1}{c}{0}                \\
FMNIST (s) Random   & \multicolumn{1}{c}{{2416}}  & \multicolumn{1}{c}{T, S, R, G}          & \multicolumn{1}{c}{AD, SGD}        & \multicolumn{1}{c}{\textbf{1e-6}} & \multicolumn{1}{c}{-}        & \multicolumn{1}{c}{1e-3, 1e-4}  & \multicolumn{1}{c}{0, 0.5}           \\
FMNIST (s) Fixed    & \multicolumn{1}{c}{{2416}}  & \multicolumn{1}{c}{T, S, R, G}          & \multicolumn{1}{c}{AD, SGD}        & \multicolumn{1}{c}{\textbf{1e-6}} & \multicolumn{1}{c}{-}        & \multicolumn{1}{c}{1e-3, 1e-4}  & \multicolumn{1}{c}{0, 0.5}           \\
CIFAR-10 (l) Seed   & \multicolumn{1}{c}{{10760}} & \multicolumn{1}{c}{G}                   & \multicolumn{1}{c}{AD}             & \multicolumn{1}{c}{\textbf{1e-5}}  & \multicolumn{1}{c}{1e-4}          & \multicolumn{1}{c}{1.00E-02}    & \multicolumn{1}{c}{0}                \\
CIFAR-10 (l) Random & \multicolumn{1}{c}{{10760}} & \multicolumn{1}{c}{T, S, R, G}          & \multicolumn{1}{c}{AD, SGD}        & \multicolumn{1}{c}{\textbf{5e-6}} & \multicolumn{1}{c}{-}        & \multicolumn{1}{c}{1e-2,1e-3}   & \multicolumn{1}{c}{0, 0.5}           \\
CIFAR-10 (l) Fixed  & \multicolumn{1}{c}{{10760}} & \multicolumn{1}{c}{T, S, R, G}          & \multicolumn{1}{c}{AD, SGD}        & \multicolumn{1}{c}{\textbf{5e-6}} & \multicolumn{1}{c}{-}        & \multicolumn{1}{c}{1e-2,1e-3}   & \multicolumn{1}{c}{0, 0.5}           \\
CIFAR-10 (s) Seed   & \multicolumn{1}{c}{2816}    & \multicolumn{1}{c}{G}                   & \multicolumn{1}{c}{AD}             & \multicolumn{1}{c}{\textbf{5e-6}} & \multicolumn{1}{c}{1e-4}          & \multicolumn{1}{c}{1e-2}    & \multicolumn{1}{c}{0}                \\
USPS (s) Seed       & \multicolumn{1}{c}{2416}    & \multicolumn{1}{c}{T}                   & \multicolumn{1}{c}{AD}             & \multicolumn{1}{c}{3e-4}          & \multicolumn{1}{c}{\textbf{1e-3}} & \multicolumn{1}{c}{1e-3}    & \multicolumn{1}{c}{0}                \\
STL-10 (l) Seed     & \multicolumn{1}{c}{10760}   & \multicolumn{1}{c}{T}                   & \multicolumn{1}{c}{AD}             & \multicolumn{1}{c}{1e-4}          & \multicolumn{1}{c}{\textbf{1e-3}} & \multicolumn{1}{c}{1e-3}    & \multicolumn{1}{c}{0}                \\
STL-10 (s) Seed     & \multicolumn{1}{c}{2816}    & \multicolumn{1}{c}{T}                   & \multicolumn{1}{c}{AD}             & \multicolumn{1}{c}{1e-4}          & \multicolumn{1}{c}{\textbf{1e-3}} & \multicolumn{1}{c}{1e-3}    & \multicolumn{1}{c}{0}                \\
\bottomrule  
\end{tabular}
\caption[Model Configuration for Sparsification]{Model Configuration for Sparsification. $\textit{Model Zoo}$ contains information about the dataset used for training, the architecture of the models in the zoo (CNN (s) - small or CNN (l) - large) and the configuration of the original model zoo (Seed, Fixed Seed or Random Seed). $\textit{W}$ denotes the number of the parameters of the models in a zoo. $\textit{Activation}$ denotes the activation function used: T - Tanh, S - Sigmoid, R - ReLU, G - GeLU. $\textit{Optim}$ denotes the optimizer: AD - Adam, SGD - Stochastic Gradient Descent. $\textit{LR VD}$ denotes the learning rate used in VD. $\textit{LR MP}$ denotes the learning rate used in magnitude pruning. $\textit{WD}$ denotes weight decay used. $\textit{Dropout}$ defines the dropout rate used.}
\label{table:twinzoooverview}
\end{table}
 
\section{Results Magnitude Pruning}
\label{AppResMP}
\begin{table}[H]
\tiny
\begin{tabular}{llllllllllllll}
\toprule
\textbf{Model Zoo}                 & \textbf{Metric} & \multicolumn{12}{c}{\textbf{Epoch}}                                                                                                                                                                                                                                                                                                                                                                              \\
\midrule
                                   &                 & \multicolumn{1}{c}{0}          & \multicolumn{1}{c}{15}         & \multicolumn{1}{c}{0}           & \multicolumn{1}{c}{15}          & \multicolumn{1}{c}{0}           & \multicolumn{1}{c}{15}         & \multicolumn{1}{c}{0}           & \multicolumn{1}{c}{15}         & \multicolumn{1}{c}{0}           & \multicolumn{1}{c}{15}          & \multicolumn{1}{c}{0}          & \multicolumn{1}{c}{15}          \\
\midrule
All                                & Spars           & \multicolumn{1}{r}{0.0 (0.0)}                            & \multicolumn{1}{r}{0.0 (0.0)}                             & \multicolumn{1}{r}{50.0 (0.0)}                             & \multicolumn{1}{r}{50.0 (0.0)}                             & \multicolumn{1}{r}{60.0 (0.0)}                             & \multicolumn{1}{r}{60.0 (0.0)}                            & \multicolumn{1}{r}{70.0 (0.0)}                             & \multicolumn{1}{r}{70.0 (0.0)}                           & \multicolumn{1}{r}{80.0 (0.0)}                             & \multicolumn{1}{r}{80.0 (0.0)}                             & \multicolumn{1}{r}{90.0 (0.0)}                           & \multicolumn{1}{r}{90.0 (0.0)}                            \\
\midrule
MNIST (s) S    & Acc             & \multicolumn{1}{r}{91.1 (0.9)} & \multicolumn{1}{r}{94.5 (0.5)} & \multicolumn{1}{r}{56.1 (13.3)} & \multicolumn{1}{r}{93.3 (0.7)}  & \multicolumn{1}{r}{41.2 (13.0)} & \multicolumn{1}{r}{92.6 (1.0)} & \multicolumn{1}{r}{28.6 (10.7)} & \multicolumn{1}{r}{91.1 (1.8)} & \multicolumn{1}{r}{19.1 (7.7)}  & \multicolumn{1}{r}{83.7 (13.5)} & \multicolumn{1}{r}{12.4 (4.0)} & \multicolumn{1}{r}{46.6 (25.1)} \\
                                   & GGap            & \multicolumn{1}{r}{0.6 (0.3)}  & \multicolumn{1}{r}{0.4 (0.2)}  & \multicolumn{1}{r}{-0.4 (0.6)}  & \multicolumn{1}{r}{0.1 (0.2)}   & \multicolumn{1}{r}{-0.2 (0.6)}  & \multicolumn{1}{r}{-0.1 (0.3)} & \multicolumn{1}{r}{-0.1 (0.5)}  & \multicolumn{1}{r}{-0.4 (0.3)} & \multicolumn{1}{r}{-0.1 (0.4)}  & \multicolumn{1}{r}{-0.7 (0.4)}  & \multicolumn{1}{r}{-0.0 (0.3)} & \multicolumn{1}{r}{-0.5 (1.0)}  \\
\midrule
MNIST (s) F   & Acc             & \multicolumn{1}{r}{75.1 (34.6)}          & \multicolumn{1}{r}{76.2 (34.4)}          & \multicolumn{1}{r}{61.8 (33.5)} & \multicolumn{1}{r}{73.7 (34.9)} & \multicolumn{1}{r}{50.0 (31.0)}           & \multicolumn{1}{r}{72.2 (35.1)}          & \multicolumn{1}{r}{35.4 (25.7)}           & \multicolumn{1}{r}{69.0 (35.1)}          & \multicolumn{1}{r}{19.7 (14.4)}           & \multicolumn{1}{r}{59.8 (34.4)}           & \multicolumn{1}{r}{11.5 (3.7)}          & \multicolumn{1}{r}{33.7 (29.2)}           \\
                                   & GGap            & \multicolumn{1}{r}{-0.1 (0.5)}          & \multicolumn{1}{r}{ -5.5 (8.1)}          & \multicolumn{1}{r}{-0.4 (0.5)}  & \multicolumn{1}{r}{-6.1 (8.5)}  & \multicolumn{1}{r}{-0.4 (0.6)}           & \multicolumn{1}{r}{-6.5 (9.0)}          & \multicolumn{1}{r}{-0.3 (0.6)}           & \multicolumn{1}{r}{-7.1 (10.2)}          & \multicolumn{1}{r}{-0.1 (0.4)}           & \multicolumn{1}{r}{-7.2 (11.4)}           & \multicolumn{1}{r}{-0.0 (0.3)}          & \multicolumn{1}{r}{-2.4 (6.3)}           \\
\midrule
MNIST (s) R    & Acc             & \multicolumn{1}{r}{74.4 (35.0)}          & \multicolumn{1}{r}{75.5 (34.8)}          & \multicolumn{1}{r}{62.0 (33.8)} & \multicolumn{1}{r}{73.3 (35.1)} & \multicolumn{1}{r}{50.3 (31.2)}           & \multicolumn{1}{r}{71.8 (35.3)}          & \multicolumn{1}{r}{35.1 (25.1)}           & \multicolumn{1}{r}{69.5 (35.1)}          & \multicolumn{1}{r}{20.1 (14.2)}           & \multicolumn{1}{r}{60.5 (34.7)}           & \multicolumn{1}{r}{11.6 (3.8)}          & \multicolumn{1}{r}{33.7 (29.2)}           \\
                                   & GGap            & \multicolumn{1}{r}{-0.1 (0.5)}          & \multicolumn{1}{r}{-5.6 (8.0)}          & \multicolumn{1}{r}{-0.4 (0.5)}  & \multicolumn{1}{r}{-5.9 (8.4)}  & \multicolumn{1}{r}{-0.4 (0.6)}           & \multicolumn{1}{r}{-6.3 (8.9)}          & \multicolumn{1}{r}{-0.3 (0.6)}           & \multicolumn{1}{r}{-7.0 (10.1)}          & \multicolumn{1}{r}{-0.1 (0.4)}           & \multicolumn{1}{r}{-7.2 (11.6)}           & \multicolumn{1}{r}{-0.0 (0.4)}          & \multicolumn{1}{r}{-2.2 (6.2)}           \\
\midrule
SVHN (s) S     & Acc             & \multicolumn{1}{r}{71.1 (8.0)} & \multicolumn{1}{r}{74.4 (8.4)} & \multicolumn{1}{r}{44.8 (8.7)}  & \multicolumn{1}{r}{72.6 (7.9)}  & \multicolumn{1}{r}{30.1 (7.4)}  & \multicolumn{1}{r}{70.7 (7.7)} & \multicolumn{1}{r}{19.0 (4.9)}  & \multicolumn{1}{r}{67.1 (7.2)} & \multicolumn{1}{r}{13.2 (3.6)}  & \multicolumn{1}{r}{58.5 (7.3)}  & \multicolumn{1}{r}{11.7 (3.8)} & \multicolumn{1}{r}{34.0 (5.7)}  \\
                                   & GGap            & \multicolumn{1}{r}{2.8 (0.7)}  & \multicolumn{1}{r}{2.8 (0.7)}  & \multicolumn{1}{r}{2.2 (1.3)}   & \multicolumn{1}{r}{2.7 (0.7)}   & \multicolumn{1}{r}{1.0 (1.2)}   & \multicolumn{1}{r}{2.5 (0.8)}  & \multicolumn{1}{r}{0.0 (0.9)}   & \multicolumn{1}{r}{2.3 (0.8)}  & \multicolumn{1}{r}{-0.4 (0.8)}  & \multicolumn{1}{r}{1.8 (0.9)}   & \multicolumn{1}{r}{-0.4 (0.8)} & \multicolumn{1}{r}{-0.1 (1.0)}  \\
\midrule
FMNIST (s) S   & Acc             & \multicolumn{1}{r}{72.7 (1.0)} & \multicolumn{1}{r}{75.5 (1.3)} & \multicolumn{1}{r}{52.1 (8.4)}  & \multicolumn{1}{r}{74.4 (1.2)}  & \multicolumn{1}{r}{43.7 (8.7)}  & \multicolumn{1}{r}{74.2 (1.2)} & \multicolumn{1}{r}{34.5 (8.6)}  & \multicolumn{1}{r}{73.6 (1.3)} & \multicolumn{1}{r}{24.2 (8.0)}  & \multicolumn{1}{r}{71.9 (2.7)}  & \multicolumn{1}{r}{14.5 (5.4)} & \multicolumn{1}{r}{57.1 (14.4)} \\
                                   & GGap            & \multicolumn{1}{r}{1.7 (0.3)}  & \multicolumn{1}{r}{2.3 (0.6)}  & \multicolumn{1}{r}{0.6 (0.4)}   & \multicolumn{1}{r}{1.9 (0.5)}   & \multicolumn{1}{r}{0.4 (0.4)}   & \multicolumn{1}{r}{1.7 (0.5)}  & \multicolumn{1}{r}{0.3 (0.4)}   & \multicolumn{1}{r}{1.5 (0.5)}  & \multicolumn{1}{r}{0.1 (0.3)}   & \multicolumn{1}{r}{1.3 (0.5)}   & \multicolumn{1}{r}{0.0 (0.2)}  & \multicolumn{1}{r}{0.8 (1.0)}   \\
\midrule
CIFAR-10 (s) S & Acc             & \multicolumn{1}{r}{48.7 (1.4)} & \multicolumn{1}{r}{49.7 (1.3)} & \multicolumn{1}{r}{39.4 (3.5)}  & \multicolumn{1}{r}{49.0 (1.3)}  & \multicolumn{1}{r}{32.8 (4.5)}  & \multicolumn{1}{r}{48.5 (1.3)} & \multicolumn{1}{r}{25.2 (4.8)}  & \multicolumn{1}{r}{47.3 (1.2)} & \multicolumn{1}{r}{17.8 (3.8)}  & \multicolumn{1}{r}{44.4 (1.1)}  & \multicolumn{1}{r}{11.6 (1.9)} & \multicolumn{1}{r}{34.5 (2.5)}  \\
                                   & GGap            & \multicolumn{1}{r}{0.7 (0.4)}  & \multicolumn{1}{r}{0.3 (0.4)}  & \multicolumn{1}{r}{0.2 (0.4)}   & \multicolumn{1}{r}{0.4 (0.4)}   & \multicolumn{1}{r}{0.0 (0.4)}   & \multicolumn{1}{r}{0.4 (0.4)}  & \multicolumn{1}{r}{-0.0 (0.4)}  & \multicolumn{1}{r}{0.3 (0.4)}  & \multicolumn{1}{r}{-0.0 (0.3)}  & \multicolumn{1}{r}{0.1 (0.4)}   & \multicolumn{1}{r}{0.0 (0.2)}  & \multicolumn{1}{r}{-0.3 (0.4)}  \\
\midrule
CIFAR-10 (l) S & Acc             & \multicolumn{1}{r}{61.5 (0.7)} & \multicolumn{1}{r}{62.4 (0.7)} & \multicolumn{1}{r}{53.2 (3.9)}  & \multicolumn{1}{r}{61.8 (0.7)}  & \multicolumn{1}{r}{43.7 (6.6)}  & \multicolumn{1}{r}{61.5 (0.6)} & \multicolumn{1}{r}{32.1 (7.4)}  & \multicolumn{1}{r}{60.8 (0.6)} & \multicolumn{1}{r}{22.2 (5.9)}  & \multicolumn{1}{r}{59.0 (0.6)}  & \multicolumn{1}{r}{15.9 (3.5)} & \multicolumn{1}{r}{51.4 (1.0)}  \\
                                   & GGap            & \multicolumn{1}{r}{2.0 (0.3)}  & \multicolumn{1}{r}{1.4 (0.4)}  & \multicolumn{1}{r}{1.2 (0.5)}   & \multicolumn{1}{r}{1.7 (0.3)}   & \multicolumn{1}{r}{0.7 (0.5)}   & \multicolumn{1}{r}{1.6 (0.3)}  & \multicolumn{1}{r}{0.2 (0.4)}  & \multicolumn{1}{r}{1.5 (0.3)}  & \multicolumn{1}{r}{0.1 (0.3)}  & \multicolumn{1}{r}{1.1 (0.4)}   & \multicolumn{1}{r}{0.0 (0.3)}  & \multicolumn{1}{r}{0.4 (0.4)}  \\
\midrule
USPS (s) S     & Acc             & \multicolumn{1}{r}{87.0 (1.7)} & \multicolumn{1}{r}{91.0 (0.8)} & \multicolumn{1}{r}{82.1 (5.0)}  & \multicolumn{1}{r}{90.1 (1.0)}  & \multicolumn{1}{r}{77.4 (7.6)}  & \multicolumn{1}{r}{89.9 (1.0)} & \multicolumn{1}{r}{70.0 (11.4)} & \multicolumn{1}{r}{89.5 (1.0)} & \multicolumn{1}{r}{54.9 (15.7)} & \multicolumn{1}{r}{88.5 (0.9)}  & \multicolumn{1}{r}{13.5 (9.8)} & \multicolumn{1}{r}{73.8 (17.3)} \\
                                   & GGap            & \multicolumn{1}{r}{5.1 (0.6)}  & \multicolumn{1}{r}{5.1 (0.6)}  & \multicolumn{1}{r}{4.9 (0.7)}   & \multicolumn{1}{r}{5.0 (0.6)}   & \multicolumn{1}{r}{4.7 (0.8)}   & \multicolumn{1}{r}{4.9 (0.6)}  & \multicolumn{1}{r}{4.3 (0.9)}   & \multicolumn{1}{r}{4.8 (0.6)}  & \multicolumn{1}{r}{3.6 (1.2)}   & \multicolumn{1}{r}{4.6 (0.5)}   & \multicolumn{1}{r}{1.2 (0.9)}  & \multicolumn{1}{r}{3.3 (1.6)}   \\
\midrule
STL-10 (s) S   & Acc             & \multicolumn{1}{r}{39.0 (1.1)} & \multicolumn{1}{r}{41.0 (1.1)} & \multicolumn{1}{r}{23.7 (4.7)}  & \multicolumn{1}{r}{41.4 (0.9)}  & \multicolumn{1}{r}{20.4 (3.8)}  & \multicolumn{1}{r}{41.3 (0.9)} & \multicolumn{1}{r}{15.3 (3.6)}  & \multicolumn{1}{r}{40.6 (0.8)} & \multicolumn{1}{r}{11.9 (2.6)}  & \multicolumn{1}{r}{36.6 (1.7)}  & \multicolumn{1}{r}{10.1 (0.6)} & \multicolumn{1}{r}{10.9 (3.5)}  \\
                                   & GGap            & \multicolumn{1}{r}{4.9 (0.8)}  & \multicolumn{1}{r}{8.6 (1.1)}  & \multicolumn{1}{r}{1.3 (0.8)}   & \multicolumn{1}{r}{5.3 (0.8)}   & \multicolumn{1}{r}{0.7 (0.7)}   & \multicolumn{1}{r}{4.2 (0.9)}  & \multicolumn{1}{r}{0.0 (0.5)}   & \multicolumn{1}{r}{2.6 (0.8)}  & \multicolumn{1}{r}{-0.2 (0.5)}  & \multicolumn{1}{r}{1.0 (0.7)}   & \multicolumn{1}{r}{-0.2 (0.3)} & \multicolumn{1}{r}{0.1 (0.3)}   \\
\midrule
STL-10 (l) S   & Acc             & \multicolumn{1}{r}{47.4 (0.9)} & \multicolumn{1}{r}{46.4 (0.9)} & \multicolumn{1}{r}{36.2 (3.2)}  & \multicolumn{1}{r}{47.4 (0.9)}  & \multicolumn{1}{r}{30.8 (3.9)}  & \multicolumn{1}{r}{48.0 (0.9)} & \multicolumn{1}{r}{25.2 (3.9)}  & \multicolumn{1}{r}{48.4 (0.9)} & \multicolumn{1}{r}{19.0 (3.6)}  & \multicolumn{1}{r}{47.9 (0.8)}  & \multicolumn{1}{r}{12.3 (2.6)} & \multicolumn{1}{r}{42.8 (1.0)}  \\
                                   & GGap            & \multicolumn{1}{r}{15.3 (1.0)} & \multicolumn{1}{r}{29.5 (1.6)} & \multicolumn{1}{r}{6.0 (1.3)}   & \multicolumn{1}{r}{20.8 (1.2)}  & \multicolumn{1}{r}{3.6 (1.1)}   & \multicolumn{1}{r}{17.6 (1.1)} & \multicolumn{1}{r}{1.9 (0.9)}   & \multicolumn{1}{r}{13.6 (0.9)} & \multicolumn{1}{r}{0.7 (0.7)}   & \multicolumn{1}{r}{8.9 (0.8)}   & \multicolumn{1}{r}{-0.1 (0.5)} & \multicolumn{1}{r}{3.5 (0.8)}   \\
\bottomrule
\end{tabular}
\caption[Magnitude Pruning Resulsts]{Magnitude Pruning Results. The results are reported as mean (std) in \%. Metric denotes the performance metric: Acc - Accuracy and GGAP - Generalization Gap. The table contains the results of all Seed, Fixed Seed and Random Seed Model Zoos before and after fine-tuning. The performance is reported for the sparsity levels 0\%, 50\%, 60\%, 70\%, 80\% and 90\%. Within the declaration of the model zoo \textit{S} represents Seed, \textit{F} represents Fixed Seed and \textit{R} represents Random Seed.}
\label{table:magnprunres}
\end{table}

\section{Results Variational Dropout}
\label{AppResVD}
\begin{table}[H]
\tiny
\begin{tabular}{llrrrrrrrrrrr}
\toprule
\textbf{Model Zoo}                 & \textbf{Metric} & \multicolumn{11}{c}{\textbf{Epoch}}                                                                                                                                                                                                                                         \\
\midrule
                                   & \textbf{}       & \multicolumn{1}{c}{0} & \multicolumn{1}{c}{2} & \multicolumn{1}{c}{4} & \multicolumn{1}{c}{6} & \multicolumn{1}{c}{8} & \multicolumn{1}{c}{10} & \multicolumn{1}{c}{13} & \multicolumn{1}{c}{16} & \multicolumn{1}{c}{19} & \multicolumn{1}{c}{22} & \multicolumn{1}{c}{25} \\
\midrule
MNIST (s) Seed    & Acc             & 91.1 (0.9)            & 91.1 (0.9)            & 90.5 (0.9)            & 89.6 (0.9)            & 88.9 (0.9)            & 88.4 (1.0)             & 87.9 (1.1)             & 87.7 (1.1)             & 87.6 (1.2)             & 87.6 (1.2)             & 87.6 (1.2)             \\
                                   & Spar            & 0.2 (0.1)             & 26.1 (1.3)            & 51.3 (1.3)            & 58.7 (1.2)            & 63.2 (1.1)            & 66.6 (1.1)             & 70.5 (1.0)             & 73.3 (1.0)             & 75.4 (1.1)             & 77.1 (1.1)             & 78.5 (1.1)             \\
                                   & GGap            & 0.6 (0.3)             & 0.3 (0.3)             & -0.6 (0.3)            & -3.2 (0.4)            & -5.0 (0.6)            & -5.4 (0.7)             & -4.9 (0.7)             & -4.6 (0.7)             & -4.2 (0.7)             & -4.0 (0.7)             & -3.9 (0.7)             \\
\midrule
MNIST (s) Fixed   & Acc             & 73.8 (35.2)           & 69.0 (35.2)           & 64.4 (34.9)           & 61.3 (34.2)           & 57.6 (33.6)           & 53.2 (33.5)            & 47.4 (33.6)            & 43.1 (32.8)            & 39.4 (32.3)            & 37.0 (31.3)            & 35.3 (30.3)            \\
                                   & Spar            & 1.5 (7.7)             & 28.2 (16.7)           & 46.7 (21.0)           & 56.6 (21.4)           & 63.8 (21.1)           & 68.8 (21.1)            & 74.0 (22.1)            & 77.4 (22.2)            & 80.3 (22.5)            & 82.3 (22.7)            & 83.4 (23.4)            \\
                                   & GGap            & -0.1 (0.5)            & -5.5 (10.0)           & -8.0 (10.8)           & -8.1 (11.2)           & -7.5 (11.2)           & -6.2 (12.1)            & -5.4 (11.9)            & -5.4 (11.4)            & -5.2 (11.0)            & -5.1 (11.0)            & -5.2 (10.5)            \\
\midrule
MNIST (s) Rand    & Acc             & 73.6 (35.3)           & 68.6 (35.3)           & 64.0 (35.0)           & 61.0 (34.3)           & 57.3 (33.7)           & 52.9 (33.7)            & 47.4 (33.6)            & 42.9 (32.8)            & 39.4 (32.1)            & 37.2 (31.2)            & 35.4 (30.2)            \\
                                   & Spar            & 1.5 (7.0)             & 28.5 (17.3)           & 46.8 (21.2)           & 56.8 (21.4)           & 64.1 (21.3)           & 69.0 (21.4)            & 74.2 (22.1)            & 77.5 (22.3)            & 80.3 (22.7)            & 82.4 (22.8)            & 83.5 (23.2)            \\
                                   & GGap            & -0.1 (0.5)            & -5.2 (10.0)           & -7.8 (10.8)           & -7.8 (11.1)           & -7.2 (11.0)           & -5.8 (12.2)            & -5.2 (11.8)            & -5.2 (11.1)            & -5.0 (10.9)            & -5.0 (11.0)            & -5.0 (10.5)            \\
\midrule
SVHN (s) Seed     & Acc             & 71.1 (8.0)            & 70.6 (8.1)            & 67.6 (7.5)            & 65.6 (7.3)            & 64.0 (7.3)            & 63.7 (7.3)             & 62.8 (7.3)             & 61.8 (7.2)             & 60.5 (7.0)             & 59.4 (6.8)             & 57.9 (6.6)             \\
                                   & Spar            & 0.2 (0.1)             & 16.7 (4.6)            & 34.2 (4.6)            & 44.3 (3.8)            & 51.6 (3.3)            & 56.1 (3.0)             & 60.7 (2.9)             & 63.9 (2.9)             & 66.1 (2.9)             & 67.9 (2.9)             & 69.5 (2.9)             \\
                                   & GGap            & 2.8 (0.7)             & 0.3 (0.8)             & -2.9 (0.9)            & -3.7 (1.0)            & -3.8 (1.1)            & -4.6 (1.2)             & -5.2 (1.2)             & -5.5 (1.3)             & -5.8 (1.4)             & -6.0 (1.5)             & -6.0 (1.5)             \\
\midrule
FMNIST (s) Seed   & Acc             & 72.7 (1.0)            & 72.8 (1.0)            & 72.1 (1.0)            & 70.9 (1.0)            & 70.2 (1.0)            & 69.6 (1.1)             & 69.4 (1.1)             & 69.3 (1.2)             & 69.3 (1.2)             & 69.3 (1.3)             & 69.4 (1.3)             \\
                                   & Spar            & 0.2 (0.1)             & 26.1 (1.2)            & 49.5 (1.3)            & 57.4 (1.2)            & 62.4 (1.1)            & 66.0 (1.0)             & 69.6 (1.0)             & 72.0 (1.0)             & 73.7 (1.0)             & 75.0 (1.0)             & 76.1 (1.1)             \\
                                   & GGap            & 1.7 (0.3)             & 1.4 (0.3)             & 0.7 (0.4)             & -1.6 (0.5)            & -3.5 (0.7)            & -3.3 (0.7)             & -2.6 (0.8)             & -2.0 (0.8)             & -1.3 (0.8)             & -1.0 (0.9)             & -1.0 (0.9)             \\
\midrule
FMNIST (s)  Fixed & Acc             & 64.2 (27.3)           & 58.4 (27.1)           & 52.3 (27.6)           & 48.7 (27.0)           & 44.9 (26.2)           & 41.1 (25.8)            & 35.8 (25.1)            & 31.8 (24.0)            & 29.2 (23.2)            & 27.3 (22.4)            & 25.8 (21.7)            \\
                                   & Spar            & 1.3 (7.3)             & 29.5 (15.3)           & 49.9 (19.3)           & 61.3 (20.7)           & 68.5 (20.5)           & 72.7 (20.4)            & 76.9 (20.5)            & 79.9 (20.9)            & 82.5 (20.9)            & 84.2 (21.1)            & 85.7 (21.0)            \\
                                   & GGap            & 1.0 (0.8)             & -2.0 (7.3)            & -2.8 (8.4)            & -2.1 (9.1)            & -0.9 (9.7)            & 0.5 (10.6)             & 1.8 (11.5)             & 2.0 (11.9)             & 1.8 (11.9)             & 1.8 (11.9)             & 1.7 (11.3)             \\
\midrule
FMNIST (s) Rand   & Acc             & 64.1 (27.4)           & 58.8 (26.9)           & 52.8 (27.3)           & 49.1 (26.7)           & 45.5 (26.1)           & 41.7 (25.7)            & 36.1 (24.9)            & 32.1 (23.9)            & 29.5 (23.1)            & 27.6 (22.3)            & 25.9 (21.6)            \\
                                   & Spar            & 1.6 (8.8)             & 29.4 (15.2)           & 49.8 (19.2)           & 61.5 (20.3)           & 68.7 (20.1)           & 73.1 (19.9)            & 77.3 (19.9)            & 80.4 (20.2)            & 82.9 (20.2)            & 84.9 (19.8)            & 86.2 (20.0)            \\
                                   & GGap            & 1.0 (0.8)             & -2.2 (7.0)            & -3.1 (8.3)            & -2.3 (9.3)            & -1.2 (9.8)            & -0.0 (10.6)            & 1.6 (11.2)             & 1.8 (11.7)             & 1.6 (11.9)             & 1.6 (11.7)             & 1.7 (11.3)             \\
\midrule
CIFAR-10 (l) Seed & Acc             & 61.5 (0.7)            & 61.7 (0.7)            & 60.1 (0.7)            & 56.7 (0.8)            & 51.5 (1.1)            & 44.4 (1.8)             & 30.7 (3.3)             & 18.2 (3.8)             & 12.6 (2.8)             & 10.7 (1.6)             & 10.2 (0.9)             \\
                                   & Spar            & 4.0 (1.4)             & 18.8 (3.2)            & 45.2 (2.1)            & 61.2 (1.4)            & 72.5 (1.0)            & 80.5 (0.7)             & 88.4 (0.5)             & 93.2 (0.4)             & 96.0 (0.3)             & 97.7 (0.3)             & 98.7 (0.2)             \\
                                   & GGap            & 2.0 (0.3)             & 1.2 (0.3)             & 1.4 (0.4)             & 1.6 (0.4)             & 1.7 (0.5)             & 2.0 (0.7)              & 2.9 (1.1)              & 2.4 (1.3)              & 0.8 (0.9)              & 0.3 (0.6)              & 0.1 (0.4)              \\
\midrule
CIFAR (l) Fixed   & Acc             & 38.8 (21.9)           & 19.6 (14.1)           & 14.0 (10.1)           & 13.3 (8.6)            & 12.5 (7.1)            & 12.0 (5.9)             & 11.4 (4.2)             & 10.9 (2.9)             & 10.7 (2.3)             & 10.5 (1.7)             & 10.2 (1.1)             \\
                                   & Spar            & 22.6 (33.0)           & 74.1 (19.3)           & 84.7 (19.6)           & 84.5 (21.7)           & 83.1 (25.4)           & 82.5 (27.7)            & 82.1 (29.6)            & 82.3 (30.5)            & 82.6 (30.5)            & 83.0 (30.5)            & 83.3 (30.6)            \\
                                   & GGap            & 1.4 (2.5)             & 6.3 (7.5)             & 0.3 (1.8)             & 0.1 (1.9)             & 0.1 (1.4)             & 0.1 (1.3)              & 0.3 (1.4)              & 0.2 (1.4)              & 0.2 (1.2)              & 0.2 (1.1)              & 0.2 (1.1)              \\
\midrule
CIFAR (l) Rand    & Acc             & 39.4 (21.9)           & 19.5 (14.1)           & 14.0 (10.2)           & 13.3 (8.7)            & 12.6 (7.4)            & 12.1 (6.1)             & 11.4 (4.2)             & 10.9 (3.0)             & 10.7 (2.4)             & 10.5 (1.8)             & 10.3 (1.3)             \\
                                   & Spar            & 23.3 (33.7)           & 74.7 (17.9)           & 85.3 (17.6)           & 85.4 (19.6)           & 84.1 (23.5)           & 83.5 (26.1)            & 83.4 (27.8)            & 83.7 (28.2)            & 84.4 (28.0)            & 84.6 (28.2)            & 84.8 (28.4)            \\
                                   & GGap            & 1.4 (2.5)             & 6.5 (7.4)             & 0.4 (1.8)             & 0.0 (1.9)             & 0.2 (1.5)             & 0.2 (1.3)              & 0.2 (1.5)              & 0.3 (1.4)              & 0.2 (1.4)              & 0.2 (1.2)              & 0.1 (0.9)              \\
\midrule
CIFAR-10 (s) Seed & Acc             & 48.7 (1.4)            & 48.9 (1.4)            & 48.1 (1.4)            & 46.0 (1.5)            & 42.7 (1.6)            & 38.0 (1.7)             & 29.6 (2.3)             & 22.4 (2.7)             & 17.0 (3.6)             & 13.1 (3.4)             & 11.2 (2.4)             \\
                                   & Spar            & 1.4 (0.4)             & 12.4 (2.6)            & 34.1 (3.2)            & 49.5 (3.0)            & 61.9 (2.6)            & 71.8 (2.1)             & 82.6 (1.5)             & 89.6 (1.1)             & 93.9 (0.8)             & 96.5 (0.8)             & 98.2 (0.7)             \\
                                   & GGap            & 0.7 (0.4)             & 0.1 (0.4)             & 0.3 (0.4)             & 0.4 (0.4)             & 0.6 (0.5)             & 0.9 (0.5)              & 1.1 (0.7)              & 1.2 (0.8)              & 1.3 (1.1)              & 0.8 (1.1)              & 0.3 (0.8)              \\
\midrule
USPS (s) Seed     & Acc             & 87.0 (1.7)            & 87.4 (1.6)            & 87.0 (1.5)            & 86.0 (1.4)            & 84.8 (1.4)            & 83.4 (1.4)             & 81.7 (1.6)             & 79.7 (2.3)             & 77.2 (3.7)             & 74.5 (4.8)             & 71.4 (6.1)             \\
                                   & Spar            & 0.6 (0.2)             & 28.5 (2.2)            & 55.9 (1.8)            & 71.2 (1.2)            & 79.4 (1.0)            & 84.9 (0.8)             & 89.7 (0.5)             & 92.3 (0.5)             & 93.5 (0.5)             & 94.2 (0.6)             & 94.5 (0.6)             \\
                                   & GGap            & 5.1 (0.6)             & 4.8 (0.6)             & 4.6 (0.5)             & 4.5 (0.6)             & 4.6 (0.6)             & 4.6 (0.8)              & 4.7 (1.2)              & 4.9 (2.0)              & 5.5 (3.4)              & 5.8 (4.5)              & 6.0 (5.7)              \\
\midrule
STL-10 (l) Seed   & Acc             & 47.4 (0.9)            & 47.7 (0.9)            & 47.7 (0.9)            & 47.1 (0.9)            & 45.1 (1.0)            & 40.2 (1.1)             & 28.0 (2.4)             & 19.4 (2.7)             & 16.3 (3.2)             & 14.8 (3.5)             & 13.9 (3.6)             \\
                                   & Spar            & 0.8 (0.1)             & 12.2 (0.5)            & 31.7 (0.8)            & 49.0 (1.0)            & 64.5 (1.2)            & 77.5 (1.1)             & 90.3 (0.6)             & 95.4 (0.3)             & 97.3 (0.2)             & 98.5 (0.2)             & 99.3 (0.2)             \\
                                   & GGap            & 15.3 (1.0)            & 13.4 (1.0)            & 12.2 (1.0)            & 9.4 (0.9)             & 6.3 (0.9)             & 4.2 (0.8)              & 2.8 (1.1)              & 1.0 (1.0)              & 0.4 (0.9)              & 0.2 (0.7)              & 0.1 (0.6)              \\
\midrule 
STL-10 (s) Seed   & Acc             & 39.0 (1.0)            & 39.1 (1.0)            & 39.4 (1.0)            & 39.3 (1.0)            & 39.1 (0.9)            & 38.5 (1.0)             & 35.9 (1.2)             & 31.4 (1.6)             & 26.4 (2.0)             & 22.0 (2.6)             & 18.9 (2.8)             \\
                                   & Spar            & 0.7 (0.2)             & 6.7 (0.6)             & 20.1 (1.0)            & 31.7 (1.3)            & 42.7 (1.4)            & 53.0 (1.5)             & 66.5 (1.3)             & 77.1 (1.2)             & 85.3 (1.1)             & 91.4 (0.9)             & 95.3 (0.6)             \\
                                   & GGap            & 5.0 (0.8)             & 4.5 (0.8)             & 4.2 (0.8)             & 3.9 (0.7)             & 3.3 (0.7)             & 2.6 (0.7)              & 1.8 (0.7)              & 1.4 (0.8)              & 1.1 (0.9)              & 1.0 (1.0)              & 0.8 (0.9)             \\
\bottomrule
\end{tabular}
\caption[Sparsification Results Overview]{Variational Dropout Results. The results are reported as mean (std) in \%. Metric denotes the performance metric: $\textit{Acc}$ - Accuracy, $\textit{Spar}$ - Sparsity and $\textit{GGAP}$ - Generalization Gap. The table contains the results of all Seed, Fixed Seed and Random Seed Model Zoos at epochs 0, 2, 4, 6, 8, 10, 13, 16, 19, 22, 25.}
\label{table:vardropres}
\end{table}

\section{Sparsity per Layer VD}
\label{AppSparLayerVD}
\begin{table}[H]
\tiny
\centering
\begin{tabular}{lrrrrrrrr}
\toprule
\textbf{Model Zoo}                 & \multicolumn{1}{c}{\textbf{Epoch}} & \multicolumn{1}{c}{\textbf{Conv 1}} & \multicolumn{1}{c}{\textbf{Conv 2}} & \multicolumn{1}{c}{\textbf{Conv 3}} & \multicolumn{1}{c}{\textbf{FC 1}} & \multicolumn{1}{c}{\textbf{FC 2}} & \multicolumn{1}{c}{\textbf{Accuracy}} & \multicolumn{1}{c}{\textbf{GGAP}} \\
\midrule
MNIST (s) Seed    & 0                                  & 0.2 (0.2)                           & 0.2 (0.2)                           & 0.4 (0.2)                           & 0.2 (0.2)                         & 0.1 (0.1)                         & 91.1 (0.9)                            & 0.6 (0.3)                         \\
                                   & 5                                  & 41.4 (10.6)                         & 55.2 (7.6)                          & 40.7 (2.3)                          & 67.6 (2.9)                        & 37.7 (4.8)                        & 90.1 (0.9)                            & -1.8 (0.4)                        \\
                                   & 10                                 & 48.9 (10.6)                         & 60.7 (7.4)                          & 62.8 (2.3)                          & 84.1 (2.5)                        & 58.9 (4.7)                        & 88.4 (1.0)                            & -5.4 (0.7)                        \\
                                   & 15                                 & 52.7 (10.0)                         & 65.1 (7.1)                          & 74.2 (2.3)                          & 90.6 (1.8)                        & 70.2 (3.2)                        & 87.8 (1.1)                            & -4.8 (0.7)                        \\
                                   & 20                                 & 55.7 (9.4)                          & 68.8 (6.8)                          & 80.0 (2.1)                          & 93.3 (1.4)                        & 75.2 (2.9)                        & 87.6 (1.2)                            & -4.2 (0.7)                        \\
                                   & 25                                 & 58.4 (8.8)                          & 71.9 (6.5)                          & 83.1 (2.1)                          & 94.7 (1.3)                        & 77.5 (3.0)                        & 87.6 (1.2)                            & -3.9 (0.7)                        \\
\midrule
SVHN (s) Seed     & 0                                  & 0.1 (0.1)                           & 0.1 (0.1)                           & 1.2 (0.3)                           & 0.3 (0.2)                         & 0.1 (0.1)                         & 71.1 (8.0)                            & 2.8 (0.7)                         \\
                                   & 5                                  & 20.0 (2.9)                          & 34.6 (5.8)                          & 19.9 (1.1)                          & 58.1 (2.9)                        & 37.0 (4.0)                        & 66.5 (7.4)                            & -3.4 (1.0)                        \\
                                   & 10                                 & 27.4 (3.8)                          & 45.2 (6.8)                          & 39.6 (1.5)                          & 82.8 (2.0)                        & 62.0 (3.7)                        & 63.7 (7.3)                            & -4.6 (1.2)                        \\
                                   & 15                                 & 31.7 (4.1)                          & 51.3 (7.2)                          & 53.3 (2.1)                          & 89.4 (1.5)                        & 72.3 (2.8)                        & 62.2 (7.2)                            & -5.5 (1.3)                        \\
                                   & 20                                 & 35.1 (4.5)                          & 55.2 (7.2)                          & 62.1 (1.7)                          & 92.6 (1.1)                        & 77.7 (3.4)                        & 60.1 (6.9)                            & -5.8 (1.4)                        \\
                                   & 25                                 & 37.1 (4.7)                          & 58.0 (7.1)                          & 69.3 (1.9)                          & 94.4 (1.0)                        & 80.8 (3.2)                        & 57.9 (6.6)                            & -6.0 (1.5)                        \\
\midrule
FMNIST (s) Seed   & 0                                  & 0.2 (0.2)                           & 0.2 (0.2)                           & 0.3 (0.2)                           & 0.2 (0.2)                         & 0.1 (0.1)                         & 72.7 (1.0)                            & 1.7 (0.3)                         \\
                                   & 5                                  & 44.2 (10.6)                         & 51.7 (7.5)                          & 47.5 (4.9)                          & 67.8 (5.5)                        & 32.1 (6.0)                        & 71.6 (1.0)                            & -0.3 (0.4)                        \\
                                   & 10                                 & 54.8 (12.0)                         & 56.1 (7.2)                          & 68.3 (4.7)                          & 87.0 (4.1)                        & 59.8 (5.4)                        & 69.6 (1.1)                            & -3.3 (0.7)                        \\
                                   & 15                                 & 57.3 (12.1)                         & 59.8 (6.8)                          & 75.6 (4.1)                          & 92.9 (2.9)                        & 74.5 (5.4)                        & 69.2 (1.2)                            & -2.0 (0.8)                        \\
                                   & 20                                 & 59.3 (12.1)                         & 62.9 (6.5)                          & 79.5 (3.6)                          & 94.8 (2.3)                        & 79.5 (5.2)                        & 69.2 (1.3)                            & -1.2 (0.9)                        \\
                                   & 25                                 & 61.1 (12.1)                         & 65.5 (6.2)                          & 81.7 (3.4)                          & 95.7 (2.0)                        & 81.5 (5.2)                        & 69.4 (1.3)                            & -1.0 (0.9)                        \\
\midrule
CIFAR-10 (l) Seed & 0                                  & 0.8 (0.3)                           & 3.0 (0.6)                           & 2.9 (0.6)                           & 10.9 (2.8)                        & 11.2 (1.8)                        & 61.5 (0.7)                            & 2.0 (0.3)                         \\
                                   & 5                                  & 15.8 (2.7)                          & 62.0 (1.9)                          & 55.1 (4.1)                          & 40.2 (5.3)                        & 30.6 (9.3)                        & 58.6 (0.8)                            & 1.6 (0.4)                         \\
                                   & 10                                 & 35.3 (4.4)                          & 88.0 (1.3)                          & 84.8 (3.5)                          & 63.3 (7.3)                        & 48.1 (15.0)                       & 44.4 (1.8)                            & 2.0 (0.7)                         \\
                                   & 15                                 & 55.9 (4.6)                          & 96.5 (0.8)                          & 96.1 (1.7)                          & 81.4 (5.3)                        & 64.2 (17.8)                       & 21.7 (3.9)                            & 2.9 (1.4)                         \\
                                   & 20                                 & 74.5 (3.6)                          & 99.1 (0.4)                          & 99.3 (0.6)                          & 93.1 (2.6)                        & 78.0 (16.8)                       & 11.7 (2.4)                            & 0.6 (0.9)                         \\
                                   & 25                                 & 88.5 (2.5)                          & 99.8 (0.2)                          & 99.9 (0.2)                          & 98.2 (0.9)                        & 87.6 (12.5)                       & 10.2 (0.9)                            & 0.1 (0.4)                         \\
\midrule
CIFAR-10 (s) Seed & 0                                  & 1.0 (0.4)                           & 1.8 (0.5)                           & 0.5 (0.2)                           & 1.6 (0.4)                         & 0.7 (0.3)                         & 48.7 (1.4)                            & 0.7 (0.4)                         \\
                                   & 5                                  & 32.6 (5.5)                          & 53.7 (5.0)                          & 15.1 (1.2)                          & 40.6 (3.9)                        & 21.4 (2.9)                        & 47.2 (1.4)                            & 0.4 (0.4)                         \\
                                   & 10                                 & 64.9 (6.1)                          & 84.7 (4.4)                          & 30.0 (1.6)                          & 69.8 (5.0)                        & 42.2 (5.1)                        & 38.0 (1.7)                            & 0.9 (0.5)                         \\
                                   & 15                                 & 87.1 (3.9)                          & 95.3 (2.2)                          & 48.0 (1.9)                          & 87.5 (3.7)                        & 63.0 (6.2)                        & 24.6 (2.5)                            & 1.2 (0.8)                         \\
                                   & 20                                 & 96.1 (2.1)                          & 98.4 (0.9)                          & 68.1 (2.0)                          & 95.7 (1.9)                        & 80.4 (5.8)                        & 15.5 (3.7)                            & 1.2 (1.0)                         \\
                                   & 25                                 & 98.9 (0.9)                          & 99.5 (0.4)                          & 84.5 (1.8)                          & 98.9 (0.7)                        & 92.1 (4.0)                        & 11.2 (2.4)                            & 0.3 (0.8)                         \\
\midrule
USPS (s)  Seed    & 0                                  & 0.5 (0.2)                           & 0.9 (0.3)                           & 0.4 (0.2)                           & 0.5 (0.2)                         & 0.1 (0.1)                         & 87.0 (1.7)                            & 5.1 (0.6)                                                           \\
                                   & 5                                  & 55.4 (13.0)                         & 80.8 (2.6)                          & 30.6 (4.2)                          & 58.2 (4.8)                        & 19.2 (4.4)                        & 86.5 (1.4)                            & 4.6 (0.5)                         \\
                                   & 10                                  & 79.6 (7.9)                          & 94.6 (1.7)                          & 62.2 (6.1)                          & 85.4 (3.9)                        & 41.1 (6.3)                        & 83.4 (1.4)                            & 4.6 (0.8)                         \\
                                   & 15                                  & 89.4 (4.8)                          & 97.4 (1.1)                          & 79.5 (5.2)                          & 92.9 (2.6)                        & 59.9 (5.6)                        & 80.4 (2.0)                            & 4.8 (1.6)                         \\
                                   & 20                                  & 92.9 (3.5)                          & 97.5 (0.9)                          & 86.1 (3.9)                          & 95.3 (1.9)                        & 71.6 (4.4)                        & 76.5 (3.8)                            & 5.4 (3.5)                         \\
                                   & 25                                  & 93.5 (3.3)                          & 96.9 (1.0)                          & 88.8 (3.2)                          & 96.0 (1.5)                        & 78.3 (3.7)                        & 71.4 (6.1)                            & 6.0 (5.7)                         \\
\midrule
STL-10 (l) Seed   & 0                                  & 0.3 (0.2)                           & 0.8 (0.3)                           & 1.1 (0.3)                           & 0.5 (0.3)                         & 0.3 (0.2)                         & 47.4 (0.9)                            & 15.3 (1.0)                        \\
                                   & 5                                   & 9.7 (1.0)                           & 37.9 (1.5)                          & 49.8 (2.5)                          & 33.4 (5.5)                        & 22.9 (10.5)                       & 47.6 (0.9)                            & 10.8 (0.9)                        \\
                                   & 10                                  & 22.6 (1.6)                          & 77.4 (1.3)                          & 88.5 (2.2)                          & 67.6 (7.0)                        & 47.7 (20.5)                       & 40.2 (1.1)                            & 4.2 (0.8)                         \\
                                   & 15                                  & 41.9 (2.2)                          & 98.0 (0.5)                          & 99.0 (0.4)                          & 91.2 (2.7)                        & 68.7 (24.7)                       & 21.3 (2.7)                            & 1.5 (1.0)                         \\
                                   & 20                                  & 69.8 (2.4)                          & 99.8 (0.1)                          & 99.8 (0.1)                          & 98.5 (0.6)                        & 82.0 (19.5)                       & 15.7 (3.3)                            & 0.3 (0.9)                         \\
                                   & 25                                 & 93.6 (1.3)                          & 99.9 (0.1)                          & 99.9 (0.1)                          & 99.6 (0.2)                        & 91.1 (11.5)                       & 13.9 (3.6)                            & 0.1 (0.6)                         \\
\midrule
STL-10 (s) Seed   & 0                                  & 0.6 (0.2)                           & 1.0 (0.3)                           & 0.3 (0.2)                           & 0.4 (0.2)                         & 0.3 (0.2)                         & 39.0 (1.0)                            & 5.0 (0.8)                         \\
                                   & 5                                  & 13.0 (1.1)                          & 34.5 (1.4)                          & 11.0 (0.9)                          & 27.9 (1.6)                        & 13.4 (1.6)                        & 39.4 (1.0)                            & 4.1 (0.7)                         \\
                                   & 10                                 & 29.0 (1.7)                          & 68.6 (1.6)                          & 23.9 (1.5)                          & 57.7 (2.2)                        & 29.1 (3.1)                        & 38.5 (1.0)                            & 2.6 (0.7)                         \\
                                   & 15                                 & 48.3 (2.1)                          & 90.0 (1.5)                          & 39.3 (1.6)                          & 81.0 (2.6)                        & 45.9 (5.5)                        & 33.1 (1.4)                            & 1.4 (0.8)                         \\
                                   & 20                                 & 74.0 (2.3)                          & 97.5 (0.9)                          & 58.5 (1.6)                          & 93.2 (2.0)                        & 63.4 (8.1)                        & 24.8 (2.2)                            & 1.1 (1.0)                         \\
                                   & 25                                 & 93.5 (1.4)                          & 99.2 (0.5)                          & 77.3 (1.4)                          & 97.5 (1.2)                        & 78.4 (8.9)                        & 18.9 (2.8)                            & 0.8 (0.9)                \\
\bottomrule
\end{tabular}
\caption[Sparsity per Layer]{Sparsity per layer at epoch 0, 5, 10, 15, 20 and 25 for all Seed model zoos. The mean (std) are reported in \%. \textit{Conv} is the abbreviation for convolutional layer, \textit{FC} is the abbreviation for fully-connected layer.}
\label{table:sparsityperlayer}
\end{table}

\section{Sparsity per Layer MP}
\label{AppSparLayerMP}
\begin{table}[H]
\centering
\tiny
\begin{tabular}{lrrrrrrrr}
\toprule
\textbf{Model   Zoo}               & \multicolumn{1}{l}{\textbf{Spars}} & \multicolumn{1}{l}{\textbf{Conv1}} & \multicolumn{1}{l}{\textbf{Conv2}} & \multicolumn{1}{l}{\textbf{Conv3}} & \multicolumn{1}{l}{\textbf{FC1}} & \multicolumn{1}{l}{\textbf{FC2}} & \multicolumn{1}{l}{\textbf{Accuracy}} & \multicolumn{1}{l}{\textbf{GGAP}} \\
\midrule
MNIST (s) Seed    & 50                                    & 42.5 (11.5)                         & 51.1 (7.6)                          & 76.4 (3.0)                          & 53.6 (2.3)                        & 25.5 (2.3)                        & 93.3 (0.7)                            & 0.1 (0.2)                         \\
                                   & 70                                    & 62.7 (14.5)                         & 72.1 (8.5)                          & 91.3 (1.8)                          & 73.8 (2.3)                        & 40.7 (3.1)                        & 91.1 (1.8)                            & -0.4 (0.3)                        \\
                                   & 90                                    & 86.6 (12.1)                         & 92.2 (5.3)                          & 98.6 (0.5)                          & 92.4 (1.4)                        & 67.1 (4.3)                        & 46.6 (25.1)                           & -0.5 (1.0)                        \\
\midrule
MNIST (s) Fixed   & 50                                    & 37.8 (4.0)                          & 63.2 (3.5)                          & 35.1 (2.2)                          & 40.6 (3.4)                        & 23.3 (3.3)                        & 73.7 (34.9)                           & -6.1 (8.5)                        \\
                                   & 70                                    & 58.0 (4.7)                          & 83.7 (2.6)                          & 51.1 (2.6)                          & 62.2 (4.0)                        & 35.9 (3.8)                        & 69.0 (35.1)                           & -7.1 (10.2)                       \\
                                   & 90                                    & 87.2 (3.1)                          & 96.6 (1.2)                          & 75.9 (2.3)                          & 90.2 (2.7)                        & 58.5 (4.7)                        & 33.7 (29.2)                           & -2.4 (6.3)                        \\
\midrule
MNIST (s) Random  & 50                                    & 37.9 (2.7)                          & 63.0 (2.6)                          & 35.4 (1.0)                          & 40.9 (2.3)                        & 24.1 (2.8)                        & 73.3 (35.1)                           & -5.9 (8.4)                        \\
                                   & 70                                    & 58.1 (2.8)                          & 83.7 (1.7)                          & 51.5 (1.2)                          & 62.4 (3.0)                        & 36.1 (3.2)                        & 69.5 (35.1)                           & -7.0 (10.1)                       \\
                                   & 90                                    & 87.4 (1.7)                          & 96.5 (0.6)                          & 76.1 (1.3)                          & 90.4 (1.9)                        & 58.9 (4.1)                        & 33.7 (29.2)                           & -2.2 (6.2)                        \\
\midrule
SVHN (s) Seed     & 50                                    & 50.0 (0.0)                          & 50.0 (0.0)                          & 50.0 (0.0)                          & 50.0 (0.0)                        & 50.0 (0.0)                        & 72.6 (7.9)                            & 2.7 (0.7)                         \\
                                   & 70                                    & 70.0 (0.0)                          & 70.0 (0.0)                          & 70.0 (0.0)                          & 70.0 (0.0)                        & 70.0 (0.0)                        & 67.1 (7.2)                            & 2.3 (0.8)                         \\
                                   & 90                                    & 90.0 (0.0)                          & 90.0 (0.0)                          & 90.0 (0.0)                          & 90.0 (0.0)                        & 90.0 (0.0)                        & 34.0 (5.7)                            & -0.1 (1.0)                        \\
\midrule
USPS (s) Seed     & 50                                    & 41.4 (9.2)                          & 64.4 (3.8)                          & 29.0 (4.4)                          & 42.0 (4.0)                        & 11.2 (2.6)                        & 90.1 (1.0)                            & 5.0 (0.6)                         \\
                                   & 70                                    & 62.9 (11.1)                         & 86.5 (2.8)                          & 45.6 (6.3)                          & 62.0 (5.0)                        & 18.4 (3.9)                        & 89.5 (1.0)                            & 4.8 (0.6)                         \\
                                   & 90                                    & 94.0 (4.4)                          & 99.5 (0.3)                          & 77.6 (7.0)                          & 89.2 (3.9)                        & 37.4 (5.7)                        & 73.8 (17.3)                           & 3.3 (1.6)                         \\
\midrule
FMNIST (s) Seed   & 50                                    & 51.8 (11.6)                         & 50.5 (7.8)                          & 67.3 (4.6)                          & 52.3 (5.0)                        & 28.5 (4.1)                        & 74.4 (1.2)                            & 1.9 (0.5)                         \\
                                   & 70                                    & 72.3 (13.2)                         & 71.1 (8.6)                          & 85.0 (3.3)                          & 72.0 (5.2)                        & 46.6 (6.5)                        & 73.6 (1.3)                            & 1.5 (0.5)                         \\
                                   & 90                                    & 91.2 (8.5)                          & 91.3 (5.1)                          & 97.1 (1.1)                          & 91.0 (3.2)                        & 73.7 (8.3)                        & 57.1 (14.4)                           & 0.8 (1.0)                         \\
\midrule
CIFAR-10 (s) Seed & 50                                    & 40.2 (5.8)                          & 59.2 (10.1)                         & 63.0 (5.2)                          & 46.2 (12.6)                       & 47.9 (4.8)                        & 49.0 (1.3)                            & 0.4 (0.4)                         \\
                                   & 70                                    & 61.2 (6.2)                          & 80.2 (9.6)                          & 84.0 (4.4)                          & 65.5 (15.5)                       & 68.1 (5.4)                        & 47.3 (1.2)                            & 0.3 (0.4)                         \\
                                   & 90                                    & 89.0 (3.5)                          & 96.5 (3.7)                          & 98.0 (1.3)                          & 86.5 (14.6)                       & 90.8 (3.4)                        & 34.5 (2.5)                            & -0.3 (0.4)                        \\
\midrule
CIFAR-10 (l) Seed & 50                                    & 13.8 (2.2)                          & 55.0 (2.0)                          & 53.0 (4.1)                          & 38.2 (1.8)                        & 20.0 (1.9)                        & 61.8 (0.7)                            & 1.7 (0.3)                         \\
                                   & 70                                    & 23.0 (3.3)                          & 76.5 (1.8)                          & 74.9 (4.4)                          & 51.6 (2.1)                        & 25.8 (2.6)                        & 60.8 (0.6)                            & 1.5 (0.3)                         \\
                                   & 90                                    & 43.4 (4.7)                          & 95.0 (0.9)                          & 95.7 (2.0)                          & 75.3 (2.2)                        & 38.6 (4.2)                        & 51.4 (1.0)                            & 0.4 (0.4)                         \\
\midrule
STL-10 (s) Seed   & 50                                    & 43.5 (1.7)                          & 66.9 (10.6)                         & 71.6 (1.5)                          & 36.1 (14.9)                       & 31.5 (1.6)                        & 41.4 (0.9)                            & 5.3 (0.8)                         \\
                                   & 70                                    & 66.5 (1.6)                          & 89.6 (10.4)                         & 94.2 (1.2)                          & 52.9 (17.5)                       & 48.1 (1.8)                        & 40.6 (0.8)                            & 2.6 (0.8)                         \\
                                   & 90                                    & 99.2 (0.3)                          & 99.9 (0.3)                          & 99.9 (0.1)                         & 80.1 (10.3)                       & 79.5 (2.2)                        & 10.9 (3.5)                            & 0.1 (0.3)                         \\
\midrule
STL-10 (l) Seed   & 50                                    & 19.7 (1.3)                          & 47.0 (1.6)                          & 63.9 (1.6)                          & 28.7 (1.4)                        & 12.4 (1.8)                        & 47.4 (0.9)                            & 20.8 (1.2)                        \\
                                   & 70                                    & 28.8 (1.5)                          & 68.2 (1.5)                          & 86.3 (1.4)                          & 41.7 (1.6)                        & 18.1 (2.6)                        & 48.4 (0.9)                            & 13.6 (0.9)                        \\
                                   & 90                                    & 44.4 (1.7)                          & 95.3 (0.7)                          & 99.2 (0.4)                          & 63.2 (1.5)                        & 28.1 (3.8)                        & 42.8 (1.0)                            & 3.5 (0.8)                         \\         
\bottomrule
\end{tabular}
\caption[Sparsity per Layer MP]{Overview of the sparsity per layer per model zoo. The values mean (std) are reported in \% for the sparsity ratios 50\%, 70\% and 90\%. \textit{Conv} is the abbreviation for convolutional layer. \textit{FC} is the abbreviation for fully-connected layer.}
\label{table:sparsitylayerMP}
\end{table}

\section{Model Agreement}
\label{AppAggr}
\begin{table}[H]
\centering
\scriptsize
\begin{tabular}{lllllll}
\toprule
\textbf{Model Zoo} & \textbf{Acc MP} & \textbf{Spar MP} & \textbf{Aggr MP}                  & \textbf{Acc VD} & \textbf{Spar VD}    & \textbf{Aggr VD}      \\
\midrule
MNIST (s) Seed     & 83.7 (13.5)         & 80.0 (0.0)            & 82.1 (13.0)             & 87.6 (1.2)      & 78.0 (1.1)           & {\textbf{83.4 (1.4)}} \\
SVHN (s) Seed      & 70.7 (7.7)           & 60.0 (0.0)            & {\textbf{74.9 (2.5)}}   & 62.2 (7.2)      & 62.8 (2.9)           & 57.5 (6.0)           \\
FMNIST (s) Seed    & 73.6 (1.3)           & 70.0 (0.0)            & {\textbf{79.7 (1.9)}}   & 69.3 (1.2)      & 72.0 (1.0)           & 76.2 (2.1)           \\
CIFAR-10 (s) Seed  & 47.3 (1.2)           & 70.0 (0.0)            & {\textbf{78.4  (1.2)}}  & 40.5 (1.5)      & 67.1 (2.8)           & 61.0 (2.7)           \\
USPS (s) Seed      & 73.8 (17.3)          & 90.0 (0.0)            & 74.3 (17.3)             & 82.3 (1.5)      & 88.5 (0.6)           & {\textbf{86.6 (1.1)}} \\
STL-10 (s) Seed    & 40.6 (0.8)           & 70.0 (0.0)            & \textbf{55.9 (2.8)}     & 35.9 (1.2)      & 66.5 (1.3)           & 54.4 (2.8)    \\
\bottomrule
\end{tabular}
\caption[Agreement of Models]{Agreement overview between original and twin Seed model zoos. The values mean (std) are reported in \%. Higher values indicate higher agreement. \textit{Aggr} denotes Agreement. \textit{Acc} denotes Accuracy. \textit{Spar} denotes Sparsity. \textit{MP} denotes Magnitude Pruning.  \textit{VD} denotes Variational Dropout.}
\label{table:agreement}
\end{table}

\section{Auto-Encoder Model}
\label{AppAE}
\begin{figure}[H]
    \centering
    \includegraphics[width=\textwidth]{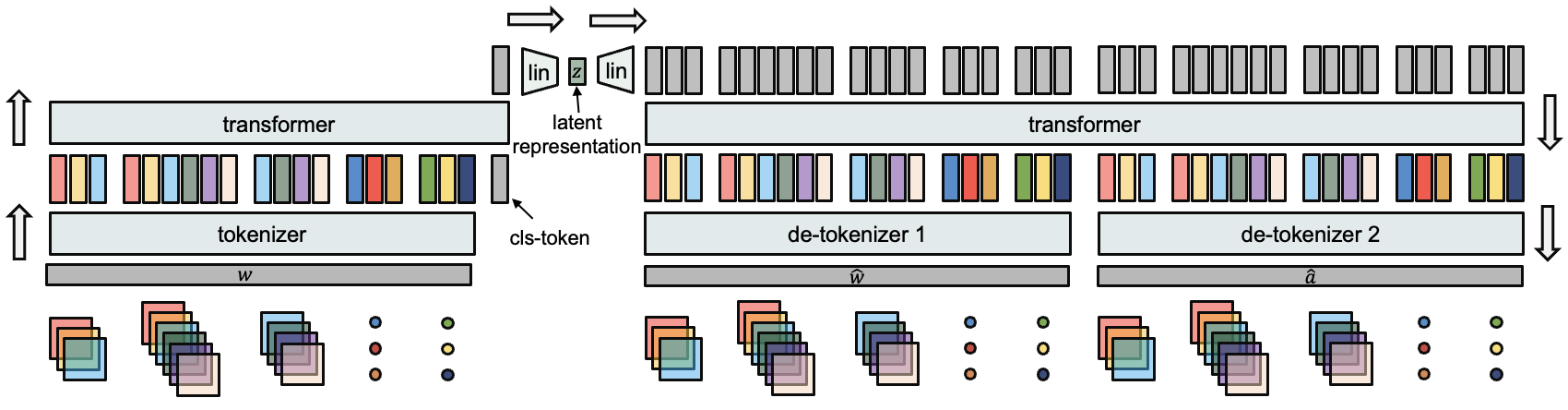}
    \caption[Auto-Encoder]{Overview of the adapted multi-head attention-based autoencoder based on \cite{schurholt2022gener}. Weights and biases are put into vector form. This sequence is converted to token embeddings, a position encoding is added and a CLS token is appended to the sequence. This sequence is passed through several layers of multi-head self-attention before the CLS token is linearly compressed by the encoder. The bottleneck $z$ is linearly decompressed by the decoder to two sequences of token embeddings. One sequence for $w$ and one sequence for $\alpha$. A position and type encoding is added, to tell the decoder if it reconstructs $w$ or $\alpha$. The two sequences are merged and and passed through another stack of multi-head self-attention. This enables the transformer to benefit from mutual information as it takes into account both sequences at once. In the end, the sequences are split again and mapped back to the original parameters.}
    \label{fig:autoencoder}
\end{figure}

\label{AppSSL}
\section{Self-Supervised Losses}
The implemented loss is composed of a contrastive part and a reconstruction part. The contrastive loss implemented is the regular NTXent loss of \cite{chen2020ntxent}. In this case each sample is randomly augmented to obtain the views $i$ and $j$. Using cosine similarity, the loss can be written as
\begin{equation} \label{eq:ntxent} 
\mathcal{L}_C = \sum_{(i,j)} -\log \frac{exp(sim(\mathbf{\bar{z}}_i,\mathbf{\bar{z}}_j))/T}{\sum_{k=1}^{2M_B} \mathbb{I}_{k\ne i} exp(sim(\mathbf{\bar{z}}_i,\mathbf{\bar{z}}_j))/T},
\end{equation} 
where $M_B$ is a batch of model parameter, $\mathbb{I}_{k\ne i}$ = 1 if $k \ne i$ and 0 otherwise and $T$ is the temperature parameter \cite{schuerholt2021ssl,chen2020ntxent}.

For the reconstruction of $w$ the layer-wise reconstruction loss of \cite{schurholt2022gener} is utilized
\begin{equation} \label{eq:reconw}
\mathcal{L}_{\bar{MSE}}^w = \frac{1}{MN}\sum_{i=1}^M \sum_{l=1}^L \Big | \Big| \frac{\mathbf{\hat{w}}_i^{(l)}-\mu_l}{\sigma_l} - \frac{\mathbf{w}_i^{(l)}-\mu_l}{\sigma_l} \Big| \Big|_2^2 = \frac{1}{MN} \sum_{i=1}^M\sum_{l=1}^L \frac{||\mathbf{\hat{w}}_i^{(l)} - \mathbf{w}_i^{(l)}||^2_2}{\sigma_l^2},
\end{equation}
where $\hat{\mathbf{w}_i^{(l)}}$ is the reconstruction of the parameters in layer $l$ of model $i$ and $\sigma_l$ and $\mu_l$ are the standard deviation and mean of the weights in layer $l$.

The novel normalization of $\alpha$ focuses more on the values around the pruning threshold of 3 and less around the values around the zero point. Accordingly the reconstruction loss for $\alpha$ is defined as
\begin{equation} \label{eq:recona_app}
\mathcal{L}_{MSE}^{\alpha} = \frac{1}{M}\sum_{i=1}^M \Big | \Big| \tanh\bigl({\frac{\mathbf{\hat{\alpha}}_i-t}{r}}\bigr) - \tanh\bigl({\frac{\mathbf{\alpha}_i-t}{r}\bigr) \Big| \Big|_2^2} ,
\end{equation}
where $t$ refers to the threshold and $r$ to the selected range of interest.

Putting the individual parts together gives the following loss:
\begin{equation} \label{eq:ssl1}
\mathcal{L}_{EcD} = \beta * \mathcal{L}_{C} + (1-\beta) * (\mathcal{L}_{\bar{MSE}}^w + \mathcal{L}_{MSE}^{\alpha}),
\end{equation}
where $\beta$ is the weighting between the contrastive and reconstruction part.


\end{document}